%% file: main.tex
\documentclass[10pt,twocolumn,letterpaper]{article}

\usepackage[accsupp]{axessibility}
\usepackage{iccv}
\usepackage{times}
\usepackage{epsfig}
\usepackage{graphicx}
\usepackage{amsmath}
\usepackage{amssymb}
\usepackage{booktabs}
\usepackage{multirow}
\usepackage{makecell}
\usepackage{float}
\usepackage{subcaption}
\usepackage{pifont}
\usepackage{threeparttable}
\usepackage[table]{xcolor}
\usepackage{arydshln}
\definecolor{lightgrey}{RGB}{244,244,244}
\definecolor{grey}{RGB}{128,128,128}
\definecolor{midgrey}{RGB}{225,225,225}
\definecolor{forestgreen}{RGB}{47, 159, 87}
\newcommand{\cmark}{\color{forestgreen}\ding{51}}%
\newcommand{\xmark}{\color{red}\ding{55}}%
\newcommand{\tablestyle}[2]{\setlength{\tabcolsep}
{#1}\renewcommand{\arraystretch}{#2}\centering\small}
\newlength\savewidth\newcommand\shline{\noalign{\global\savewidth\arrayrulewidth
  \global\arrayrulewidth 1pt}\hline\noalign{\global\arrayrulewidth\savewidth}}

\usepackage{colortbl}  
\usepackage{array}
\definecolor{Gray}{gray}{0.94}
\definecolor{liGray}{gray}{0.5}
\definecolor{LightCyan}{rgb}{0.88,1,1}

\usepackage[pagebackref=true,breaklinks=true,letterpaper=true,colorlinks,bookmarks=false]{hyperref}

\iccvfinalcopy 

\def\iccvPaperID{1714} 

\ificcvfinal\fi

\begin{document}

\title{Disentangling Spatial and Temporal Learning for Efficient \\ Image-to-Video Transfer Learning}
\vspace{-0.8cm}

\author{Zhiwu Qing$^{1}$ \quad Shiwei Zhang$^{2*}$ \quad Ziyuan Huang$^{3}$ \quad Yingya Zhang$^2$  \\ Changxin Gao$^{1}$ \quad Deli Zhao$^2$    \quad Nong Sang$^{1*}$ \\
$^1$Key Laboratory of Image Processing and Intelligent Control \\ School of Artificial Intelligence and Automation, Huazhong University of Science and Technology \\ 
$^2$Alibaba Group \quad $^3$ARC, National University of Singapore \\ 
{\tt\small \{qzw, cgao, nsang\}@hust.edu.cn \quad \{zhangjin.zsw, yingya.zyy\}@alibaba-inc.com} \\
{\tt\small ziyuan.huang@u.nus.edu \quad zhaodeli@gmail.com} 
\vspace{-0.6cm}
}





\maketitle
\let\thefootnote\relax\footnotetext{$^*$Corresponding authors.}
\ificcvfinal\thispagestyle{empty}\fi

\begin{abstract}
\vspace{-3mm}
%
Recently, large-scale pre-trained language-image models like CLIP have shown extraordinary capabilities for understanding spatial contents, but naively transferring such models to video recognition still suffers from unsatisfactory temporal modeling capabilities.
%
%
Existing methods insert tunable structures into or in parallel with the pre-trained model, which either requires back-propagation through the whole pre-trained model and is thus resource-demanding, or is limited by the temporal reasoning capability of the pre-trained structure. 
In this work, we present DiST, which \underline{di}sentangles the learning of \underline{s}patial and \underline{t}emporal aspects of videos. 
Specifically, DiST uses a dual-encoder structure, where a pre-trained foundation model acts as the spatial encoder, and a lightweight network is introduced as the temporal encoder. 
An integration branch is inserted between the encoders to fuse spatio-temporal information. 
The disentangled spatial and temporal learning in DiST is highly efficient because it avoids the back-propagation of massive pre-trained parameters. 
Meanwhile, we empirically show that disentangled learning with an extra network for integration benefits both spatial and temporal understanding. 
Extensive experiments on five benchmarks show that DiST delivers better performance than existing state-of-the-art methods by convincing gaps.
When pre-training on the large-scale Kinetics-710, we achieve 89.7\% on Kinetics-400 with a frozen ViT-L model, which verifies the scalability of DiST.
Codes and models can be found in \href{https://github.com/alibaba-mmai-research/DiST}{\textcolor{blue}{https://github.com/alibaba-mmai-research/DiST}}.

\end{abstract}

\vspace{-4mm}

\section{Introduction}
\label{sec:intro}
\begin{figure}
    \centering
    \includegraphics[width=1.0\linewidth]{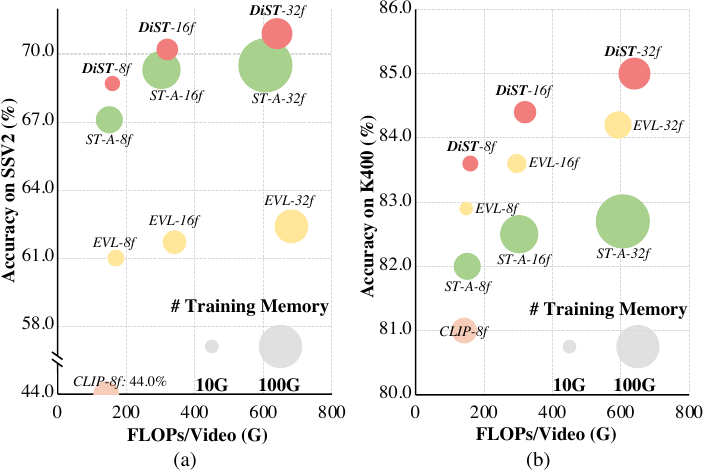}
    \vspace{-7.5mm}
    \caption{
    Accuracy vs. per-video GFLOPs on  SSV2~\cite{goyal2017ssv2} and K400~\cite{kay2017k400} with ViT-B/16~\cite{dosovitskiy2020-vit}.
    ``EVL''~\cite{lin2022evl}: Efficient Video Learning.
    ``ST-A''~\cite{pan2022st-adapter}: ST-Adapter.
    ``CLIP'': Fully fine-tuning the CLIP pre-trained image encoder.
    }
    \label{fig:first_fig}
    \vspace{-0.5cm}
\end{figure}

\begin{figure*}
\centering
\centering
\begin{minipage}{0.74\linewidth}{\begin{center}
    \includegraphics[width=1.0\linewidth]{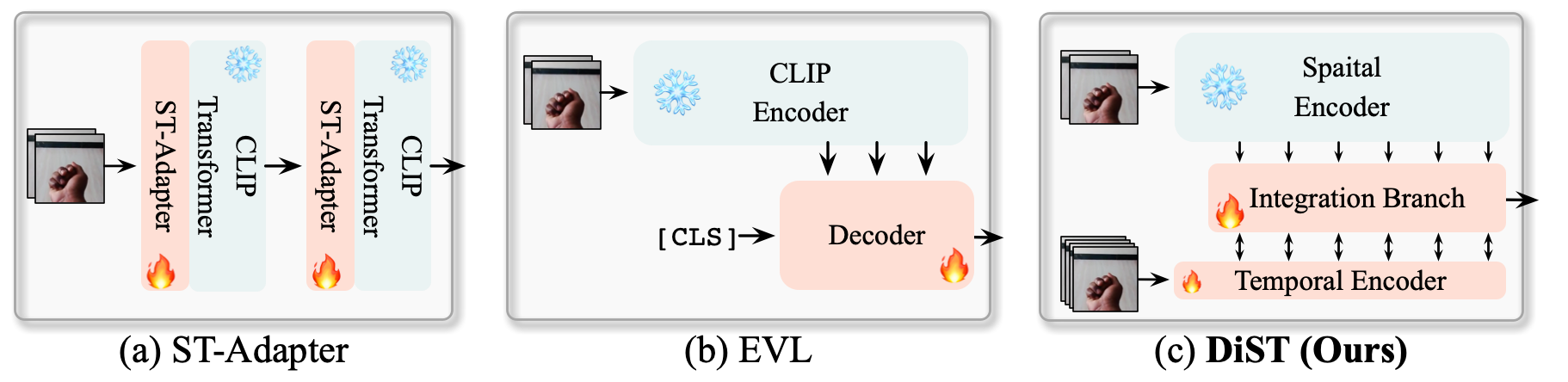}
\vspace{-7mm}
\end{center}}
\end{minipage}
\hspace{-2mm}
\begin{minipage}{0.25\linewidth}{\begin{center}
\tablestyle{1.2pt}{1.4}
\vspace{-8mm}
    \begin{tabular}{cc|ccc}
         &Method  & \makecell{BP-\\Free}  & Temp. & Spat.\\
        \shline
        (a)  &\scriptsize \makecell{ST-Adapter} & \xmark  & {\color{forestgreen}Strong} & {\color{forestgreen}Strong} \\
        (b)  & \makecell{EVL} & \cmark  & {\color{red}Weak} & {\color{forestgreen}Strong} \\
        (c)  & \textbf{DiST} & \cmark  & {\color{forestgreen}Strong} & {\color{forestgreen}Strong} \\
    \end{tabular}
\vspace{-6mm}
\end{center}}
\end{minipage}
\caption{Comparison with existing efficient fine-tuning approaches for video recognition. 
\textbf{(a)} ST-Adapter~\cite{pan2022st-adapter}.
\textbf{(b)} EVL~\cite{lin2022evl}.
\textbf{(c)} Our proposed DiST.
``BP-Free'' indicates ``back-propagation-free'' for the encoder.
``Temp.'' and ``Spat.'' are ``temporal modeling'' and ``spatial modeling'', respectively.
}
\vspace{-5mm}
\label{fig:framework-compare}
\end{figure*}


Video understanding is a fundamental yet challenging research topic in computer vision. 
Early approaches for this task learn spatio-temporal representations by designing different architectures, such as two-stream models~\cite{karpathy2014twostream, wang2016tsn}, 3D networks~\cite{tran2015c3d, feichtenhofer2019slowfast, kay2017k400}, Transformers~\cite{arnab2021vivit, bertasius2021timesformer, yan2022multiview}, and they have achieved impressive progress on some challenging benchmarks~\cite{kay2017k400, goyal2017ssv2}.
Recently, a new paradigm that transfers the large-scale language-image pre-training models, \eg,  CLIP~\cite{radford2021clip}, to video understanding tasks~\cite{ju2022promptingclipvid,lin2022evl,pan2022st-adapter,li2022uniformerv2} has been drawing lots of attention due to its remarkable spatial modeling potential, and it deserves the enhancement of the potential for spatio-temporal reasoning.
%
%

As in Fig.~\ref{fig:framework-compare}~(a), a popular design for efficient transfer learning is to insert tunable structures between pre-trained Transformer blocks~\cite{pan2022st-adapter,lian2022ssfada,gao2021clipadapter,chen2022adaptformer}. 
Parameter-efficient as it is, it would require back-propagation through massive parameters that are supposed to be frozen, which is inefficient in training. 
With a large number of video frames, this inefficiency hinders the scaling of large video-text foundation models under limited GPU memory, as in Fig.~\ref{fig:first_fig}.
To tackle this, the recent work~\cite{lin2022evl} introduces a decoder in parallel with the pre-trained encoder. 
%
This indeed increases training efficiency by avoiding back-propagation through pre-trained parameters. 
However, the major function of the decoder in such an approach is to collect relevant information from the frozen encoder, which makes the output of the decoder highly correlated to the spatial information provided by the pre-trained image Transformer.

In this work, we present DiST, a dual-encoder framework for efficiently transferring the pre-trained image-text foundation models to video-text ones. 
DiST shares the merits of both frameworks mentioned above by disentangling the spatial and temporal modeling: 
\textbf{(i)} by connecting all the structures in parallel to the frozen model, DiST avoids the back-propagation through the massive parameters in the pre-trained Transformer;
\textbf{(ii)} by introducing a separated encoder that specifically designed to extract temporal information for the input video, the temporal modeling capability is enhanced. 
Further, to simultaneously exploit the spatial semantics and the temporal information extracted by the dual-encoder structure, an integration branch is imposed to fuse the features from both spatial and temporal encoders.

We evaluate DiST on three challenging supervised video recognition benchmarks, \ie, Kinetics-400~\cite{kay2017k400}, Something-Something v2~\cite{goyal2017ssv2}, Epic-Kitchens-100~\cite{damen2020ek100}, and two zero-shot benchmarks, \ie, UCF101~\cite{soomro2012ucf101} and HMDB51~\cite{jhuang2011hmdb51}.
Our DiST achieves state-of-the-art performance on all datasets with convincing gains to existing approaches.
Moreover, under limited resources, DiST enables us to pre-train on large-scale video datasets since only the lightweight temporal encoder and integration branch require pre-training.
With Kinetics-710 for pre-training, we verify the superior scalability of DiST and achieve better performance than fully fine-tuned ones.


\section{Related Works}
\label{sec:related_works}


\begin{figure*}
    \centering
    \includegraphics[width=0.9\linewidth]{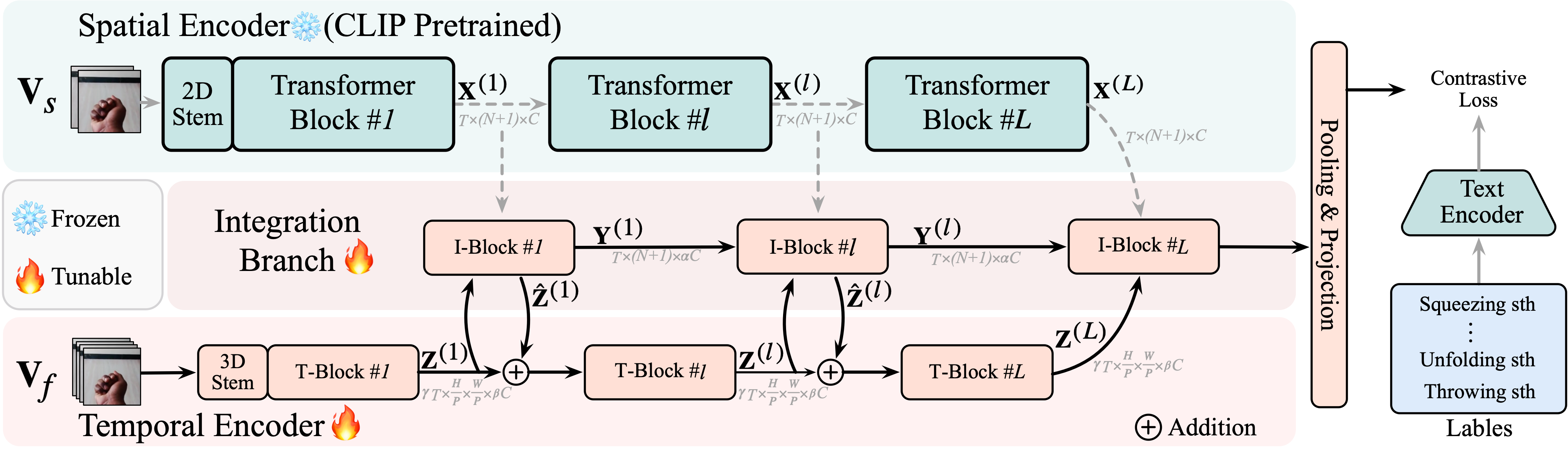}
    \caption{The overall framework of DiST. DiST contains three components, including a spatial encoder, a temporal encoder and an integration branch. The spatial encoder is a CLIP~\cite{radford2021clip} pre-trained Vision Transformer (ViT)~\cite{dosovitskiy2020-vit}, which is frozen and back-propagation-free in training. The temporal encoder is composed of a series of lightweight temporal blocks (T-Block), which is responsible for capturing dynamic temporal patters in videos. The integration branch consists of multiple interaction blocks (I-Block), which are designed to integrate the features from the spatial encoder and the temporal encoder.}
    \label{fig:framewrok}
    \vspace{-4mm}
\end{figure*}

\noindent
\textbf{Visual-language Pre-training.}
Recently, visual language pre-training~\cite{miech2019howto100m,miech2020milnce,sun2019videobert,zhu2020actbert,li2022flip,radford2021clip,yuan2021florence,jia2021scaling,yu2022coca} has made remarkable progress. 
One of the most representative works is CLIP~\cite{radford2021clip}. 
Following that, a series of prompt-based~\cite{li2021prefix,zhou2022coop,zhou2022cocop,bahng2022visualprompt,zhang2021vt-clip} and adapter-based~\cite{pfeiffer2020adapterfusion,gao2021clipadapter,lian2022ssfada,sung2022vl-adapter} works explored how to efficiently transfer the pre-trained models to image tasks.
Meanwhile, transferring language-image pre-trained models to videos~\cite{wang2021actionclip,pan2022st-adapter,lin2022evl,ni2022expanding-xclip,ju2022prompting,cheng2021improving} has attracted wide attention due to its striking performance.
For example, Ni \textit{et al.}~\cite{ni2022expanding-xclip} proposed adopting a cross-frame attention module and video-specific text prompts for remarkable video ``zero-shot'' generalization ability. 
EVL~\cite{lin2022evl}  employed frozen CLIP models to extract video features for efficient video learning. 
Pan \textit{et al.}~\cite{pan2022st-adapter} inserted spatial-temporal adapters (ST-Adapter) into pre-trained transformer blocks to enable space-time modeling capabilities in image models.

EVL~\cite{lin2022evl} is mostly related to ours. 
It uses a transformer decoder to collect spatial information from frozen features, which is also back-propagation-free for pre-trained parameters. 
However, our DiST adopts a dual encoder structure to exploit video specific  temporal changes, and enjoys both strong space-time modeling and high training efficiency.

\noindent
\textbf{Video Recognition.}
One of the key aspects of video recognition is exploring temporal patterns in videos. 
For convolutional-based methods~\cite{simonyan2014twostream,tran2015c3d,qiu2017p3d,wang2018tsn,carreira2017i3d,wang2018artnet,tran2019csn,feichtenhofer2019slowfast,tran2018r21d,lin2019tsm,yang2020tpn,wang2021tdn,huang2021tada,feichtenhofer2020x3d}, which introduce 3D convolutions~\cite{tran2015c3d,carreira2017i3d}, factorized spatial and temporal convolutions~\cite{qiu2017p3d,tran2018r21d,feichtenhofer2019slowfast}, and convolutional modules with temporal modeling capabilities~\cite{jiang2019stm,lin2019tsm,qiu2019learning,li2020tea,wang2020video,huang2021tada}. 
Due to the limited receptive field of convolutional networks, Transformer-based approaches~\cite{arnab2021vivit,bertasius2021timesformer,fan2021mvit,li2022mvitv2,ryoo2021tokenlearner,li2022uniformer,liu2022videoswin,patrick2021motionformer,neimark2021vtn,bulat2021xvit} with global attention have achieved promising performance. For example, ViViT~\cite{arnab2021vivit} and Transformer~\cite{liu2022videoswin} achieve space-time modeling by factorized spatial and temporal transformers and window attention, respectively. 
Apart from designing the model architecture, recently, self-supervised video representation learning~\cite{pan2021videomoco,feichtenhofer2021largescale,qian2021cvrl,jenni2021time,diba2021vi2clr,feichtenhofermasked,tong2022videomaenju,wei2022maskedfeat,wang2022bevt} has also gained popularity due to its impressive performance.

Our multi-branch design shares similar spirit with SlowFast~\cite{feichtenhofer2019slowfast} and Multiview Transformers~\cite{yan2022multiview}, which both design different views for similar network structures, and all parameters require back-propagation for training.
Nevertheless, our work designs an asymmetric network structure with strong temporal modeling capability, and gets rid of back-propagation for massive parameters.

\section{Approach}

\label{sec:approach}
In this work, we seek to empower the large-scale pre-trained language-image models with spatial-temporal modeling capability in training efficient way.
Specifically, as shown in Fig.~\ref{fig:framewrok}, the proposed DiST comprises three components: the spatial encoder, temporal encoder, and integration branch.
The spatial encoder is a heavy CLIP pre-trained Vision Transformer (ViT), which extracts frozen features for sparse frames with powerful spatial semantics.
The temporal encoder is a lightweight spatio-temporal network with low-channel capacity, adopting dense frames as input and capturing temporal patterns specific to video understanding.
The integration branch links the spatial and temporal encoders by interacting with the disentangled spatial and temporal features.
In this section, we first briefly present the formulation of the spatial encoder in Sec.~\ref{sec:approach_spat}. Then, the temporal encoder and integration branch are elaborated in  Sec.~\ref{sec:approach_temporal} and Sec.~\ref{sec:approach_inter}, respectively. Finally, the training loss is introduced in Sec.~\ref{sec:approach_pooling_loss}.

\subsection{Spatial Encoder}
\label{sec:approach_spat}

%
In DiST, the spatial encoder is an off-the-shelf feature extractor without recording the gradients, resulting in significant efficiency improvements during training.
%
It extracts independent spatial features from several sparse frames. 
%
%
Given a video clip $\mathbf{V}_s\in \mathbb{R}^{T\times H\times W \times 3}$, where $T$, $H$ and $W$ are the frame number, height, and width, respectively. 
%
%
Following ViT~\cite{dosovitskiy2020-vit}, each frame is split into $N=\frac{H}{P}\times\frac{W}{P}$ patches, and the patch size is denoted as $P\times P$. 
Then, these small patches are projected by a fully connected layer, \ie, the 2D stem in Fig.~\ref{fig:framewrok}, which generates a sequence of patch embeddings $[\mathbf{x}^{(0)}_{t, 1}, \mathbf{x}^{(0)}_{t, 2}, \cdots, \mathbf{x}^{(0)}_{t, N}]$, where $t=\{1, \cdots, T\}$ is the frame index.
Next, an additional learnable token $\mathbf{x}_{\text{cls}}$ is concatenated for each frame, 
and the full inputs of Transformer blocks are denoted as:
\begin{equation}
    \mathbf{X}^{(0)}_t=[\mathbf{x}^{(0)}_{t, \text{cls}}, \mathbf{x}^{(0)}_{t, 1}, \mathbf{x}^{(0)}_{t, 2}, \cdots, \mathbf{x}^{(0)}_{t, N}] + \mathbf{e}^{\text{spatial}}, 
    \label{eq:seq_0}
\end{equation}
where the $(N+1)$ embeddings are enhanced with the trainable spatial position embedding $\mathbf{e}^{\text{spatial}}$.
Assuming that the spatial encoder has $L$ Transformer blocks, the features of the $l_{\text{th}}$ layer for the $t_{\text{th}}$ frame can be extracted by:
\begin{equation}
    \mathbf{X}_t^{(l)}=\text{Transformer}^{(l)}(\mathbf{X}_t^{(l-1)}) \in \mathbb{R}^{(N+1)\times C},
\end{equation}
where $l=\{1,\cdots,L\}$ refers to the layer index and $C$ is the channel dimension.
We adopt the notation $\mathbf{X}^{(l)}=[\mathbf{X}_1^{(l)},\cdots,\mathbf{X}_T^{(l)}]\in \mathbb{R}^{T\times(N+1)\times C}$ to represent the spatial features of $T$ frames in the $l_{\text{th}}$ layer.

\begin{figure}
    \centering
    \includegraphics[width=1.0\linewidth]
    {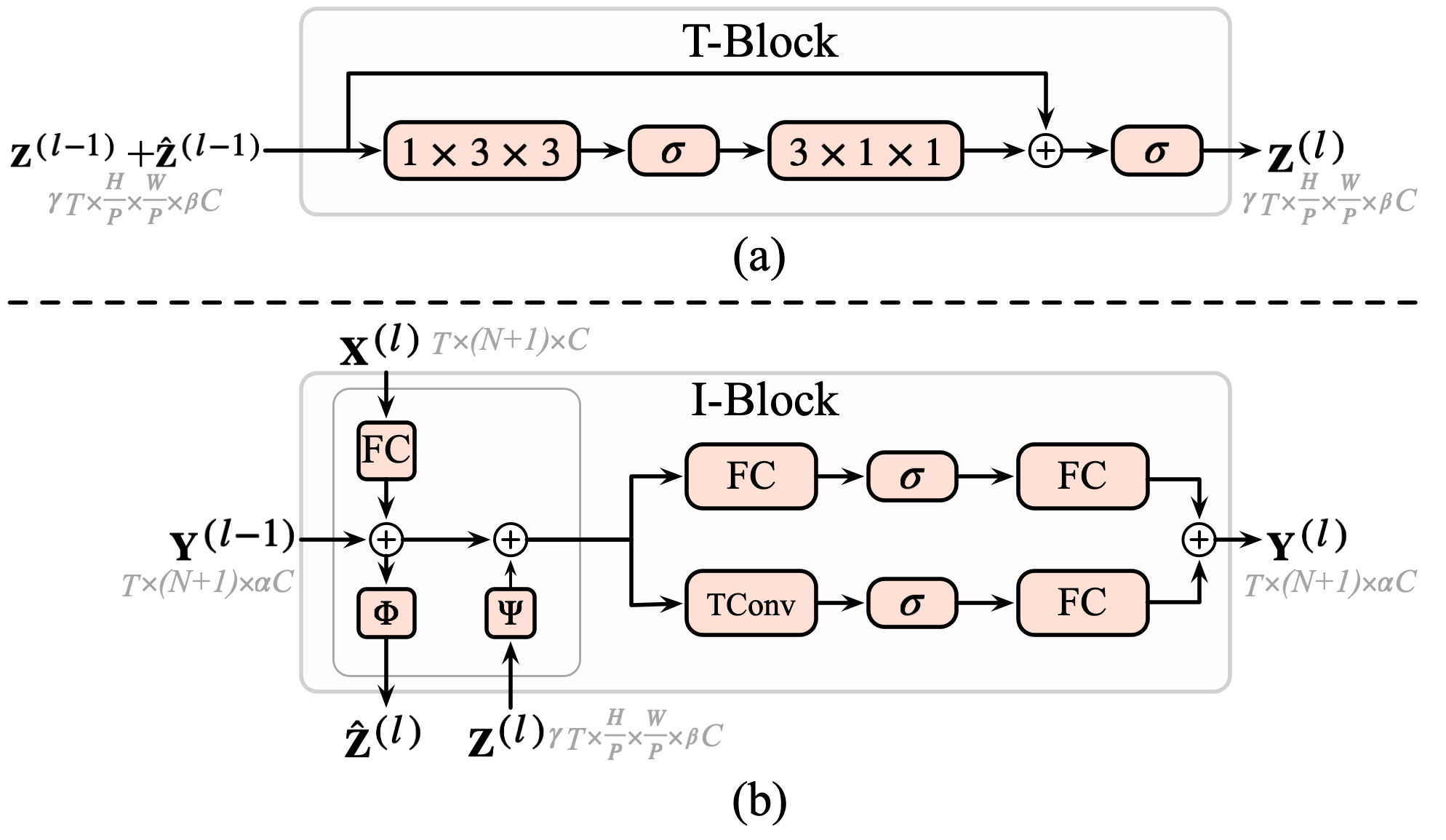}
    \vspace{-8mm}
    \caption{\textbf{(a)} shows the structural details of Temporal Block (T-Block). Our default structure is R(2+1)D~\cite{tran2018r21d}.
    \textbf{(b)} illustrates the structural details of Interaction Block (I-Block).}
    \label{fig:st-block-tffn}
    \vspace{-4mm}
\end{figure}

\subsection{Temporal Encoder}
\label{sec:approach_temporal}

With powerful spatial semantics from the CLIP pre-trained spatial encoder, we expect to train a lightweight temporal-specific network to disentangle the fine-grained motion learning for video understanding.
Therefore, we design a general temporal encoder, which can utilize the original video frames as input and receive the semantic guidance from the spatial encoder.
Without losing generality, we assume that the number of frames in the temporal encoder is $\gamma T$, $\gamma\in\{1, 2, 4\}$, which means that the temporal input $\mathbf{V}_f\in \mathbb{R}^{\gamma T\times H\times W \times 3}$ can be sampled around the spatial input $\mathbf{V}_s$ by $\gamma$ times.
%
Next, $\mathbf{V}_f$ is projected by a 3D convolution, \ie, the 3D stem in Fig.~\ref{fig:framewrok}, for patch embedding. The kernel size and stride of the 3D stem in spatial dimension are both $P$ to align with the spatial size in the spatial encoder for feature integration. 
Thus the projected temporal features can be formulated as: $\mathbf{Z}^{(0)}=\text{Conv3d}(\mathbf{V}_f)\in \mathbb{R}^{\gamma T \times \frac{H}{P} \times \frac{W}{P}\times \beta C}$. Here, $\beta\in\{\frac{1}{24}, \frac{1}{12}, \frac{1}{6}, \frac{1}{4}\}$, indicates the channel reduction rate of the temporal encoder. 
%
Note that we do not perform temporal downsampling to preserve more temporal details.
Then, a series of lightweight Temporal Blocks (T-Block) are designed to extract spatio-temporal patterns for these frames, which can be written as:
\begin{equation}
\label{eq:tblock_phi}
    \mathbf{Z}^{(l)}=\text{T-Block}^{(l)}(\mathbf{Z}^{(l-1)} + \mathbf{\hat{Z}}^{(l-1)}) \in \mathbb{R}^{\gamma T \times \frac{H}{P} \times \frac{W}{P}\times \beta C},
\end{equation}
where the function $\text{T-Block}(\cdot)$ is the smallest unit that can perform spatio-temporal modeling. 
As shown in Fig.~\ref{fig:st-block-tffn}~(a), the classic R(2+1)D~\cite{tran2018r21d} is adopted for $\text{T-Block}(\cdot)$ by default. 
Meanwhile, other optional designs, such as the convolution-based C3D~\cite{tran2015c3d}, TAdaConv~\cite{huang2021tada} and the transformer-based joint space-time transformer~\cite{tong2022videomaenju}, have also been explored in Tab.~\ref{tab:st_cell}.
Although the joint transformer generates a large self-attention matrices, it still feasible due to the low channel capacity of the temporal encoder.
%
The $\mathbf{\hat{Z}}^{(l-1)}$ indicates the interaction features from the integration branch. 
It will be introduced in the next section.


\subsection{Integration Branch}
\label{sec:approach_inter}

The role of the integration branch can be summarized into two aspects: 
\textit{(i)} receiving and integrating the spatial and temporal features into more discriminative spatio-temporal representations; 
\textit{(ii)} performing feature interactions between the spatial encoder and temporal encoder. 
Specifically, it transfers the powerful semantics from the spatial encoder to temporal encoder, thus guiding the random initialized temporal encoder to capture temporal clues.

The integration branch is composed of a series of Interaction Blocks (I-Block).
The structural details of one I-Block are shown in Fig.~\ref{fig:st-block-tffn}.
Formally, for $\mathbf{X}^{(l)}$ from the spatial encoder and $\mathbf{Z}^{(l)}$ from the temporal encoder, we first adopt addition to absorb them into the integrated features $\mathbf{Y}^{(l-1)}\in \mathbb{R}^{T\times (N+1)\times \alpha C}$ from the previous layer, and then perform spatio-temporal fusion by a temporal Feed Forward Network.
This can be expressed as:
\begin{small}
\begin{equation}
\label{eq:tffn}
\begin{split}
    \mathbf{\hat{Y}}^{(l)} &= \mathbf{Y}^{(l-1)} + \text{FC}(\mathbf{X}^{(l)}) + \Psi^{(l)}(\mathbf{Z}^{(l)}),\\
    \mathbf{Y}^{(l)} &= \text{FC}(\sigma(\text{FC}(\text{LN}(\mathbf{\hat{Y}}^{(l)})))) + 
    \text{FC}(\sigma(\text{TConv}(\text{LN}(\mathbf{\hat{Y}}^{(l)})))).
\end{split}
\end{equation}
\end{small}

\noindent
Here, $\mathbf{Y}^{(0)}$ is 0. $\text{FC}(\mathbf{X}^{(l)})$ reduces the channel dimension from $C$ to $\alpha C$. $\Psi^{(l)}(\cdot)$ is the lateral interaction from the temporal encoder to the integration branch that will be discussed later. $\text{FC}(\cdot)$, $\text{LN}(\cdot)$, $\text{TConv}(\cdot)$ and $\sigma(\cdot)$ are abbreviations of the linear layer, layer normalization, 1D convolution with the kernel size of $3\times1\times1$ and activation function, respectively. The 1D convolution introduced here is to further encourage the spatio-temporal blending for the disentangled spatial and temporal features.

Based on the elaborately designed architecture mentioned above, the integration branch is capable of simultaneously receiving the spatial semantics and temporal patterns, and then integrating them into unified spatio-temporal representations for video recognition.

Next, we discuss the interaction details between the integration branch and the temporal encoder.
First, the interaction function $\Psi(\cdot)$ is responsible for transmitting the information from the temporal encoder (\ie, $\mathbf{Z}^{(l)}$) to the integration branch.
In implementation, to align with the feature size of the integration branch, $\Psi(\cdot)$ needs to downsample the temporal dimension of $\mathbf{Z}^{(l)}\in \mathbb{R}^{\gamma T\times \frac{H}{P}\times\frac{W}{P}\times \beta C}$ from $\gamma T$ to $T$, then increases the channels from $\beta C$ to $\alpha C$, and appends a new class token to align the number of tokens (\ie, $N+1$) in  $\mathbf{Y}^{(l-1)}$. 
Formally, $\Psi(\cdot)$ can be written as:
\begin{equation}
\label{eq:def_psi}
    \Psi(\mathbf{Z})=[\text{Flatten}(\text{DConv}(\mathbf{Z})), \mathbf{z}_{\text{cls}}]  \in \mathbb{R}^{T\times (N+1)\times \alpha C},
\end{equation}
where $\text{DConv}(\cdot)$ is a temporal convolution for temporal downsampling, whose kernel size and stride are both $\gamma$ in temporal dimension. 
The input and output channels of $\text{DConv}(\cdot)$ are $\beta C$ and $\alpha C$. 
The function $\text{Flatten}(\cdot)$  flattens the spatial dimension $\frac{H}{P}\times\frac{W}{P}$ to $N$.
$\mathbf{z}_{\text{cls}}\in\mathbb{R}^{T \times 1\times \alpha C}$ is a trainable class token.
$[\cdot,\cdot]$ indicates the concatenation.
In this way, the function $\Psi(\cdot)$ realizes the feature alignment of the temporal branch and the integration branch.

The interaction from the integration branch to the temporal encoder is defined in Eq.~\ref{eq:tblock_phi}, \ie, the notation $\mathbf{\hat{Z}}^{(l-1)}$. For simplicity, we use $\mathbf{\hat{Z}}^{(l)}$ instead of $\mathbf{\hat{Z}}^{(l-1)}$ for discussion.
Therefore, $\mathbf{\hat{Z}}^{(l)}$ can be formulated as:
\begin{equation}
\mathbf{\hat{Z}}^{(l)}=\Phi(\mathbf{Y}^{(l-1)} + \text{FC}(\mathbf{X}^{(l)})).
\end{equation}
Here, $\mathbf{Y}^{(l-1)}$ is the integrated feature  from the previous layer and $\mathbf{X}^{(l)}$ is the spatial feature from the current layer. 
For $\mathbf{\hat{Z}}^{(0)}$, we set it to 0.
This design can ensure that the spatio-temporal semantic guidance can be timely injected into the temporal encoder.

For the function $\Phi(\cdot)$, which is responsible for the compatibility of the integration features ($\mathbb{R}^{T\times (N+1)\times \alpha C}$) with the temporal features ($\mathbb{R}^{\gamma T\times \frac{H}{P}\times\frac{W}{P}\times \beta C}$). 
%
Therefore, we first remove the spatial class token, and reduce the channels from $\alpha C$ to $\beta C$ by a linear layer. 
Then, to upsample the temporal dimension from $T$ to $\gamma T$, we adopt the nearest interpolation by default.
Finally, the $N$ tokens of each frame are reshaped to $\frac{H}{P}\times\frac{W}{P}$ to align with the temporal features in Eq.~\ref{eq:tblock_phi}.

\begin{table*}[ht]
\centering
\subfloat[
``Temp.'' is the abbreviation of ``temporal''. ``Integ.'' is the abbreviation of ``Integration''. 
\label{tab:branches}
]{
\centering
\begin{minipage}{0.31\linewidth}{\begin{center}
\tablestyle{2pt}{1.05}
    \begin{tabular}{cc|cc}
       \small \makecell{Temp.Encoder}   &\small \makecell{Integ.Branch}  &  SSV2 & K400\\
        \shline
          \multicolumn{2}{c|}{EVL~\cite{lin2022evl}}    & 61.0  & 82.9 \\
        \xmark  & \xmark & 55.0  & 79.9 \\
        \cmark  & \xmark & 63.2  & 81.8 \\
        \xmark  & \cmark & 65.9  & 83.0  \\
        \cellcolor{midgrey}\cmark  & \cellcolor{midgrey}\cmark & \cellcolor{midgrey}\textbf{68.7}  &\cellcolor{midgrey}\textbf{83.6}\\
    \end{tabular}
\end{center}}
\end{minipage}
}
\hspace{0.6em}
\subfloat[
``Integ.$\rightarrow$Temp.'' indicates the interactions from the integration branch to the temporal encoder, and vice versa.
\label{tab:feature_interaction}
]{
\centering
\begin{minipage}{0.31\linewidth}{\begin{center}
\tablestyle{2pt}{1.05}
    \begin{tabular}{cc|cc}
       \small Integ.$\rightarrow$Temp.& \small Temp.$\rightarrow$Integ. &  SSV2 & K400\\
        \shline
        \xmark  & \xmark &  67.5 &  83.0 \\
        \cmark  & \xmark &  67.9 &  83.4 \\
        \xmark  & \cmark &  67.7 &  83.1 \\
        \cellcolor{midgrey}\cmark  & \cellcolor{midgrey}\cmark & \cellcolor{midgrey}\textbf{68.7}  & \cellcolor{midgrey}\textbf{83.6}\\
        \multicolumn{4}{c}{}\\
    \end{tabular}
\end{center}}
\end{minipage}
}
\hspace{0.6em}
\subfloat[Optional designs of T-Block. ``J. Trans.'' is the space-time joint transformer.
\label{tab:st_cell}
]{
\centering
\begin{minipage}{0.31\linewidth}{\begin{center}
\tablestyle{2pt}{1.05}
    \begin{tabular}{c|ccc}
        case  &  SSV2  &K400 & GFLOPs\\
        \shline
       TAda~\cite{huang2021tada} & 67.8 &82.4& 162.0 \\
       C3D~\cite{tran2015c3d} & 67.8 &82.6& 168.2\\
       J. Trans.~\cite{tong2022videomaenju}  &  67.6 &83.5&165.0\\
       \cellcolor{midgrey}R(2+1)D~\cite{tran2018r21d} &\cellcolor{midgrey}\textbf{68.7}  & \cellcolor{midgrey}\textbf{83.6}&\cellcolor{midgrey}\textbf{163.1}\\
      \multicolumn{4}{c}{}\\
    \end{tabular}
\end{center}}
\end{minipage}
}

\subfloat[
Varying values of $\gamma$, \ie, the number of frames in the temporal encoder.
\label{tab:temporal_frames}
]{
\centering
\begin{minipage}{0.31\linewidth}{\begin{center}
\tablestyle{2pt}{1.05}
    \begin{tabular}{ccc|ccc}
        Spat.  & Temp. & $\gamma$  &  SSV2 & K400 & GFLOPs\\
        \shline
        \multirow{3}{*}{8f}  & 8f &  1 &   67.9 &83.4 & \textbf{158.7}\\
        ~  & \cellcolor{midgrey}16f &  \cellcolor{midgrey}2 &  \cellcolor{midgrey}68.7 & \cellcolor{midgrey}\textbf{83.6}&\cellcolor{midgrey}163.1\\
        ~  & 32f &  4 & \textbf{69.1} & 83.3&171.6\\
        ~  & 64f &  8 & 68.5 & 83.6 &188.8\\
    \end{tabular}
\end{center}}
\end{minipage}
}
\hspace{0.6em}
\subfloat[
Varying values of $\alpha$, \ie, the channel capacity of the integration branch.
\label{tab:integra_dim}
]{
\centering
\begin{minipage}{0.31\linewidth}{\begin{center}
\tablestyle{2pt}{1.05}
    \begin{tabular}{cc|ccc}
        Dim & $\alpha$  &  SSV2 & K400 & GFLOPs\\
        \shline
        96 &   $1/8$ & 62.6 & 79.0 & \textbf{149.4}\\
        192   &   $1/4$ &  66.7&82.0 &152.8 \\
        \cellcolor{midgrey}384   &   \cellcolor{midgrey}$1/2$ & \cellcolor{midgrey}\textbf{68.7} & \cellcolor{midgrey}\textbf{83.6}&\cellcolor{midgrey}163.1\\
         768  &   $1$ & 68.7  &83.3 &196.1\\
    \end{tabular}
\end{center}}
\end{minipage}
}
\hspace{0.6em}
\subfloat[
Varying values of $\beta$, \ie, the channel capacity of the temporal encoder.
\label{tab:temporal_dim}
]{
\centering
\begin{minipage}{0.31\linewidth}{\begin{center}
\tablestyle{2pt}{1.05}
    \begin{tabular}{cc|ccc}
        Dim & $\beta$  &  SSV2 & K400 & GFLOPs\\
        \shline
        64 &   ${1}/{12}$ & 67.9  &83.2 &\textbf{159.7}\\
        \cellcolor{midgrey}96 &  \cellcolor{midgrey}${1}/{8}$ &  \cellcolor{midgrey}68.7 & \cellcolor{midgrey}\textbf{83.6 }&\cellcolor{midgrey}163.1\\
        128 &  ${1}/{6}$ &  68.6 & 83.3&167.4\\
        192 &  ${1}/{4}$ &  \textbf{68.9}& 83.4 &178.7\\
    \end{tabular}
\end{center}}
\end{minipage}
}
\vspace{-3mm}
\caption{Ablations on \textbf{Something-Something V2} and \textbf{Kinetics-400}. Our spatial encoder is a 8-frame vanilla ViT-B/16 pre-trained by CLIP~\cite{radford2021clip} with a channel width of 768. The TSN~\cite{wang2016tsn} uniform sampling is performed on both datasets. The inference protocol of all models and datasets are 3 clips $\times$ 1 center crop.}
\vspace{-3mm}
\label{tab:ablations}
\end{table*}

Unless particularly emphasized, the above-discussed downsampling and upsampling methods are the default implementation for $\Psi(\cdot)$ and $\Phi(\cdot)$, respectively. 
There are also other alternative implementations, such as the temporal convolution in function $\Psi(\cdot)$ can be replaced with a combination of a pooling layer and a linear layer, and the nearest interpolation in $\Phi(\cdot)$ can be replaced with trilinear interpolation or deconvolution.
These optional designs will be further explored in experiments, \ie, Sec.~\ref{sec:experiments}.

\subsection{Training Loss}
\label{sec:approach_pooling_loss}
The integration of the disentangled spatial and temporal information yields semantically rich spatio-temporal representations, which can lead to more promising video recognition performances.
Next, following CLIP~\cite{radford2021clip}, we first perform adaptive pooling to obtain a video-level class token $\mathbf{y}_{\text{cls}}$ for the representation of $\mathbf{Y}^{(L)}$.
Then, the text features of the correct category labels are taken as positives, and contrastive loss is employed to train both the temporal encoder and integration branch. The formulation can be written as:
\begin{equation}
\begin{split}
    \mathbf{y}_{\text{cls}}&=\text{Proj}(\text{AdaPooling}(\mathbf{Y}^{(L)})), \\
    \mathcal{L}_{\text{CL}}&=-\text{log}\frac{\text{exp}(\text{sim}(\mathbf{y}_{\text{cls}}, \mathbf{u}_{i})/\tau)}{\sum_{k=1}^{M}\text{exp}(\text{sim}(\mathbf{y}_{\text{cls}}, \mathbf{u}_{k})/\tau)},
\end{split}
\end{equation}
where $\text{Proj}(\cdot)$ indicates the projection to the classification space.
$\text{sim}(\cdot, \cdot)$ is the normalized cosine similarity. $\mathbf{u}_{i}$ is a text feature for the $i_{\text{th}}$ label. Here, we assume the correct label of $\mathbf{y}_{\text{cls}}$ is $i$. 
$\tau$ refers to the temperature parameter.
In this manner, the proposed structure can project videos into a text space, which not only enables the video recognition but also retains the zero-shot ability for videos.

\section{Experiments}
\label{sec:experiments}

\subsection{Implementation}

\noindent
\textbf{Datasets.}
We evaluate our proposed DiST on five widely used benchmarks, \ie, Kinetics-400 (K400)~\cite{kay2017k400}, Something-Something V2 (SSV2)~\cite{goyal2017ssv2}, Epic-Kitchens-100 (EK100)~\cite{damen2020ek100}, HMDB51~\cite{jhuang2011hmdb51}, and UCF101~\cite{soomro2012ucf101}. K400 is a large scale action recognition dataset 
spanning 400 different human actions. 
SSv2 
is a commonly used temporally-heavy dataset. 
EK100 is a egocentric recorded interaction between persons and objects in the kitchen.
Each video is labeled with a verb and a noun.
UCF101 and HMDB51 are two relatively small action recognition datasets, 
which are employed for zero-shot evaluation  following~\cite{ni2022expanding-xclip}.

\noindent
\textbf{Architecture.}
Following previous work~\cite{lin2022evl}, we use the CLIP~\cite{radford2021clip} pre-trained ViT-B/16, ViT-L/14 and ViT-L/14-336p as our spatial encoder.
Unless otherwise specified, we mark the default settings in the temporal encoder and the integration branch in \colorbox{midgrey}{gray} in Sec.~\ref{sec:ablations}.

\noindent
\textbf{Training settings.}
All training and testing settings are provided in Appendix. 

\subsection{Ablation Studies}
\label{sec:ablations}



\noindent
\textbf{The role of the temporal encoder and integration branch.}
In Tab.~\ref{tab:branches}, we attempt to remove the temporal encoder and integration from DiST to observe their effects.
Compared with the spatial encoder only (the 1$st$ line), imposing temporal encoder and the integration branch  can both significantly boost performance, for example, the improvements on SSV2 reach 10.9\% and 8.2\%, respectively. 
DiST without integration branch cannot interact and integrate the independent spatial and temporal information, resulting in poorer performance.
However, with the integration branch, incorporating the temporal encoder to learn more video-specific features further yields a gain of 2.8\%, which proves the necessity of the disentangled temporal encoder.

\noindent
\textbf{The feature interactions between the temporal encoder and integration branch} are explored in Tab.~\ref{tab:feature_interaction}. 
One can observe that both directions of information transmission can improve performances, and the combination of the two can boost accuracy more significantly, \eg, the improvement can reach 1.2\% on SSV2. 
This demonstrates that the spatio-temporal blending in the integration branch and the spatial semantic guidance in temporal encoder are both essential.

\noindent
\textbf{Optional designs of T-Block.} The T-Block is intended to empower the lightweight temporal encoder with temporal modeling capabilities. 
Here, we attempt three different modules in Tab.~\ref{tab:st_cell}, \ie, the convolution-based TAdaConv~\cite{huang2021tada}, C3D~\cite{tran2015c3d} and R(2+1)D~\cite{tran2018r21d}, the transformer-based joint spatial-temporal Transformer~\cite{tong2022videomaenju}. 
The R(2+1)D outperforms the other approaches by about 1\% with less computation on SSV2.
We speculate that the lighter R(2+1)D may be easier to optimize.

\begin{figure}[t]
    \includegraphics[width=1.0\linewidth]{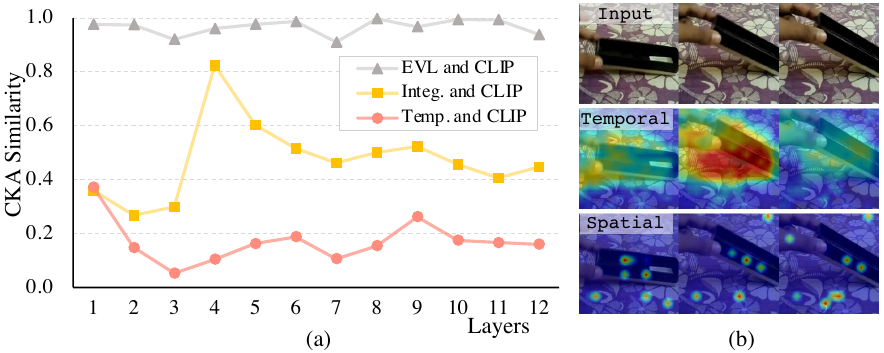}
    \vspace{-6mm}
    \caption{\textbf{(a)} We evaluate three different pairs of feature correlations by CKA similarities~\cite{cka}: \textit{(i)} EVL~\cite{lin2022evl}  features and CLIP features; \textit{(ii)} our integrated features and CLIP features; \textit{(iii)} our temporal features CLIP features. \textbf{(b)} Visualization of the magnitude of features. Red indicates a large magnitude of feature activation values, while blue indicates a small magnitude.}
    \label{fig:cka_vis_param}
    \vspace{-5mm}
\end{figure}

\noindent
\textbf{The parameter analysis of $\gamma$, $\alpha$ and $\beta$.} \textit{(i)} Our temporal encoder can receive flexible frames as input. $\gamma$ determines the number of frames input to the temporal encoder. 
Tab.~\ref{tab:temporal_frames} shows that more frames can introduce richer temporal clues and consistently boost the accuracies, especially for the temporally-heavy dataset (\ie, SSV2).
However, for computation efficiency, we set $\gamma$ to 2 by default, which can produce 0.8\% gain on SSV2 compared with $\gamma=1$.
\textit{(ii)} $\alpha$ determines the channel width of the integration branch. 
As in Tab.~\ref{tab:integra_dim}, when channel width is 96 ($\alpha=1/8$), compared with 384 ($\alpha=1/2$), the performance degradation is 6.1\% (68.7\% vs. 62.6\%) on SSV2.
This is because the integration branch requires a larger channel width to accommodate the rich semantics in spatio-temporal fusion.
Nevertheless, the channel width of 768 ($\alpha=1$) can not further improve accuracy, which means that 384 is sufficient.
\textit{(iii)} Tab.~\ref{tab:temporal_dim} explores the impact of the channel dimension of the temporal encoder. 
Since the spatial encoder provides powerful spatial semantics, the temporal encoder only needs to capture specific motions in videos.
When the channel dimension is 96 ($\beta=1/8$), it can achieve satisfactory performance, the gain is 0.8\% compared with smaller 64 dimensions.
Meanwhile, more channels are also saturated, which implies designing a lightweight temporal encoder is reasonable.

\begin{table}[t]
\centering
\hspace{1.2em}
\subfloat[
\label{tab:psi}
]{
\centering
\begin{minipage}{0.45\linewidth}{\begin{center}
    \begin{tabular}{c|c}
        Downsampling  &  Top-1\\
        \shline
       Avg Pooling & 68.3\\
       Max Pooling & 68.4\\
       \cellcolor{midgrey}DConv &  \cellcolor{midgrey}\textbf{68.7}\\
    \end{tabular}
\end{center}}
\end{minipage}
}
\subfloat[
\label{tab:phi}
]{
\centering
\begin{minipage}{0.45\linewidth}{\begin{center}
    \begin{tabular}{c|c}
        Upsampling  &  Top-1\\
        \shline
       DeConv & 67.8\\
       Trilinear &68.3\\
       \cellcolor{midgrey}Nearest & \cellcolor{midgrey}\textbf{68.7} \\
    \end{tabular}
\end{center}}
\end{minipage}
}
\vspace{-3mm}
\caption{Alternative designs for feature interactions on SSV2.
\textbf{(a)} Replacing the downsampling function in $\Psi(\cdot)$ with different pooling functions.
\textbf{(b)} Replacing the nearest interpolation in $\Phi(\cdot)$ with different upsampling functions. ``DeConv'' is deconvolution.}
    \vspace{-4mm}
\label{tab:optional_design}
\end{table}

\begin{table}[t]
    \centering
    \small
    \begin{tabular}{c|ccc|c}
        \multirow{2}{*}{Method} & \multicolumn{3}{c|}{Training}& Inference \\ 
         & Memory & Step & Epoch  & Throughput\\
        \shline
       ST-Adapter~\cite{pan2022st-adapter} & 39.10G & 0.95s & 38 & \textbf{50}\\
       EVL~\cite{lin2022evl} & 15.05G& 0.71s & 45& 38\\
       \cellcolor{midgrey}\textbf{DiST} & \cellcolor{midgrey}\textbf{12.68}G & \cellcolor{midgrey}\textbf{0.52}s & \cellcolor{midgrey}\textbf{36} & \cellcolor{midgrey}48\\
    \end{tabular}
    \caption{{Training and inference costs on SSV2.} ``Step'': the training step time with batch size of 32 on A100-80G. ``Throughput'': inference throughput (Videos/s).}
    \vspace{-4mm}
    
    \label{tab:train_infer_compare}
\end{table}

\begin{table*}[t]
    \centering
    \tablestyle{4pt}{1.05}
\begin{tabular}{ccccccccc}
Method &\makecell{Pre-train} &  Architecture & Input Size & \makecell{FLOPs$\times$Cr.$\times$Cl. (T)} &  \makecell{Param (M)}& Frozen&  \makecell{Top-1} & \makecell{Top-5}\\
\midrule[1.15pt]
SlowFast~\cite{feichtenhofer2019slowfast} & ImageNet-21K  & R101+NL & $16\times224^2$ & $0.1\times3\times1$ & 60& \xmark  & 63.1 & 87.6 \\
ViViT FE~\cite{arnab2021vivit} & IN21K+K400  & ViT-L & $16\times224^2$ & $1.0\times3\times4$& 612& \xmark  & 65.4 & 89.8 \\
MTV-B(320p)~\cite{yan2022multiview}& IN21K+K400 & - &$32\times224^2$ & $0.9\times3\times4$& 310& \xmark & 68.5 & 90.4 \\
MViT~\cite{fan2021mvit} & Kinetics-600  & MViT-B-24 & $32\times224^2$ & $0.2\times3\times1$ & 53& \xmark  & 68.7 & 91.5 \\
Video Swin~\cite{liu2022videoswin} & IN21K+K400  & Swin-B & $32\times224^2$ & $0.3\times3\times1$ & 60& \xmark & 69.6 & 92.7 \\
TAdaConvNeXtV2~\cite{huang2023tadav2} & IN1K+K400  & ConvNeXt-S & $32\times224^2$ & $0.2\times3\times2$ &  82& \xmark & 70.0 & 92.0 \\
\midrule
EVL\includegraphics[width=0.25cm,height=0.25cm]{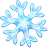}~\cite{lin2022evl} & CLIP-400M  & ViT-B & $32\times224^2$ & $0.68\times1\times3$  & 175& \cmark & 62.4 & - \\
ST-Adapter\includegraphics[width=0.25cm,height=0.25cm]{fig/frozen.pdf}~\cite{pan2022st-adapter} & CLIP-400M  & ViT-B & $32\times224^2$ & $0.61\times1\times3$ & 93& \cmark & 69.5 & \textbf{92.6} \\
\cellcolor{lightgrey}\textbf{DiST}$_{\gamma\text{=2}}$\includegraphics[width=0.25cm,height=0.25cm]{fig/frozen.pdf} &\cellcolor{lightgrey}CLIP-400M&\cellcolor{lightgrey}ViT-B &\cellcolor{lightgrey}$8\times224^2$ &\cellcolor{lightgrey}$0.16\times1\times3$ &\cellcolor{lightgrey}105&\cellcolor{lightgrey}\cmark&\cellcolor{lightgrey}68.7&\cellcolor{lightgrey}91.1\\
\cellcolor{lightgrey}\textbf{DiST}$_{\gamma\text{=2}}$\includegraphics[width=0.25cm,height=0.25cm]{fig/frozen.pdf} &\cellcolor{lightgrey}CLIP-400M&\cellcolor{lightgrey}ViT-B &\cellcolor{lightgrey}$16\times224^2$ &\cellcolor{lightgrey}$0.32\times1\times3$ &\cellcolor{lightgrey}105&\cellcolor{lightgrey}\cmark&\cellcolor{lightgrey}70.2&\cellcolor{lightgrey}92.0\\
\cellcolor{lightgrey}\textbf{DiST}$_{\gamma\text{=2}}$\includegraphics[width=0.25cm,height=0.25cm]{fig/frozen.pdf} &\cellcolor{lightgrey}CLIP-400M&\cellcolor{lightgrey}ViT-B &\cellcolor{lightgrey}$32\times224^2$ &\cellcolor{lightgrey}$0.65\times1\times3$ &\cellcolor{lightgrey}105&\cellcolor{lightgrey}\cmark&\cellcolor{lightgrey}\textbf{70.9}&\cellcolor{lightgrey}92.1\\
\midrule
\textcolor{grey}{UnifromerV2}~\cite{li2022uniformerv2} & \textcolor{grey}{CLIP-400M}  & \textcolor{grey}{ViT-L} & \textcolor{grey}{$32\times224^2$} & \textcolor{grey}{$1.73\times1\times3$}  & \textcolor{grey}{574}&\textcolor{grey}{\xmark} & \textcolor{grey}{73.0} & \textcolor{grey}{94.5} \\
\textcolor{grey}{TAdaFormer}~\cite{huang2023tadav2} & \textcolor{grey}{CLIP-400M}  & \textcolor{grey}{ViT-L} & \textcolor{grey}{$32\times224^2$} & \textcolor{grey}{$1.70\times2\times3$}  & \textcolor{grey}{364}&\textcolor{grey}{\xmark} & \textcolor{grey}{73.6} & \textcolor{grey}{-} \\
EVL\includegraphics[width=0.25cm,height=0.25cm]{fig/frozen.pdf}~\cite{lin2022evl} & CLIP-400M  & ViT-L & $32\times224^2$ & $3.21\times1\times3$ & 654& \cmark  & 66.7 &  -\\
EVL\includegraphics[width=0.25cm,height=0.25cm]{fig/frozen.pdf}~\cite{lin2022evl} & CLIP-400M  & ViT-L & $32\times336^2$ & $8.08\times1\times3$ & 654 & \cmark & 68.0 & - \\
ST-Adapter\includegraphics[width=0.25cm,height=0.25cm]{fig/frozen.pdf}~\cite{pan2022st-adapter} & CLIP-400M  & ViT-L & $32\times224^2$ & $2.75\times1\times3$& 347 & \cmark  & 72.3 & 93.9 \\
\cellcolor{lightgrey}\textbf{DiST}$_{\gamma\text{=2}}$\includegraphics[width=0.25cm,height=0.25cm]{fig/frozen.pdf} &\cellcolor{lightgrey}CLIP-400M&\cellcolor{lightgrey}ViT-L &\cellcolor{lightgrey}$8\times224^2$ &\cellcolor{lightgrey}$0.71\times1\times3$&\cellcolor{lightgrey}336& \cellcolor{lightgrey}\cmark&\cellcolor{lightgrey}70.8&\cellcolor{lightgrey}92.3\\
\cellcolor{lightgrey}\textbf{DiST}$_{\gamma\text{=2}}$\includegraphics[width=0.25cm,height=0.25cm]{fig/frozen.pdf} &\cellcolor{lightgrey}CLIP-400M&\cellcolor{lightgrey}ViT-L &\cellcolor{lightgrey}$16\times224^2$ &\cellcolor{lightgrey}$1.42\times1\times3$ &\cellcolor{lightgrey}336&\cellcolor{lightgrey}\cmark&\cellcolor{lightgrey}72.5&\cellcolor{lightgrey}93.0\\
\cellcolor{lightgrey}\textbf{DiST}$_{\gamma\text{=2}}$\includegraphics[width=0.25cm,height=0.25cm]{fig/frozen.pdf} &\cellcolor{lightgrey}CLIP-400M&\cellcolor{lightgrey}ViT-L &\cellcolor{lightgrey}$32\times224^2$ &\cellcolor{lightgrey}$2.83\times1\times3$ &\cellcolor{lightgrey}336&\cellcolor{lightgrey}\cmark&\cellcolor{lightgrey}\textbf{73.1}&\cellcolor{lightgrey}\textbf{93.2}\\

\end{tabular}
    \caption{Comparison with the state-of-the-art methods on Something-Something V2. ``Cr.'' and ``Cl.'' are the abbreviation for ``spatial crops'' and ``temporal clips''. ``Frozen'' indicates freezing the CLIP pre-trained parameters.}
    \label{tab:sota_ssv2}
\end{table*}

\begin{table*}[t]
    \centering
    \subfloat[
    \label{tab:zero_shot}
    ]{
    \centering
    \begin{minipage}{0.48\linewidth}{\begin{center}
    \tablestyle{2pt}{1.05}
    \begin{tabular}{ccc|cc}
        Method  & Model & Frames &  HMDB51 & UCF101\\
        \shline
       ActionCLIP~\cite{wang2021actionclip} & B/16 & 32$\times$1$\times$1 & 40.8$\pm$5.4 &58.3$\pm$3.4 \\
       X-CLIP~\cite{ni2022expanding-xclip} & B/16 &32$\times$1$\times$1 & 44.6$\pm$5.2 &72.0$\pm$2.3 \\
       \cellcolor{lightgrey}\textbf{DiST}$_{\gamma\text{=2}}$\includegraphics[width=0.25cm,height=0.25cm]{fig/frozen.pdf} &  \cellcolor{lightgrey}B/16 & 32$\times$1$\times$1\cellcolor{lightgrey}&\cellcolor{lightgrey}55.4$\pm$1.2 & \cellcolor{lightgrey}72.3$\pm$0.6 \\
       \cellcolor{lightgrey}\textbf{DiST}$_{\gamma\text{=2}}$\includegraphics[width=0.25cm,height=0.25cm]{fig/frozen.pdf} &  \cellcolor{lightgrey}L/14  &\cellcolor{lightgrey}32$\times$1$\times$1&\cellcolor{lightgrey}\textbf{57.5}$\pm$1.6&\cellcolor{lightgrey}\textbf{74.9}$\pm$0.8 \\
    \end{tabular}
    \end{center}}
    \end{minipage}
    }
    \subfloat[
    \label{tab:ek100}
    ]{
    \centering
    \begin{minipage}{0.48\linewidth}{\begin{center}
    \tablestyle{2pt}{1.05}
    \begin{tabular}{ccc|ccc}
        Method  & Model &  Frames & Verb & Noun & Action\\
        \shline
        EVL\includegraphics[width=0.25cm,height=0.25cm]{fig/frozen.pdf}~\cite{lin2022evl} & B/16   & 8$\times$3$\times$1 & 62.7 & 51.0 & 37.7 \\
       ST-Adapter\includegraphics[width=0.25cm,height=0.25cm]{fig/frozen.pdf}~\cite{wang2021actionclip} & B/16  & 8$\times$3$\times$1 & 67.6 & 55.0 & - \\
       \cellcolor{lightgrey}\textbf{DiST}$_{\gamma\text{=2}}$\includegraphics[width=0.25cm,height=0.25cm]{fig/frozen.pdf} &  \cellcolor{lightgrey}B/16  & \cellcolor{lightgrey}8$\times$3$\times$1&\cellcolor{lightgrey}69.5 & \cellcolor{lightgrey}58.1 & \cellcolor{lightgrey}45.8\\
       \cellcolor{lightgrey}\textbf{DiST}$_{\gamma\text{=2}}$\includegraphics[width=0.25cm,height=0.25cm]{fig/frozen.pdf} &  \cellcolor{lightgrey}L/14  & \cellcolor{lightgrey}8$\times$3$\times$1&\cellcolor{lightgrey}\textbf{70.7} & \cellcolor{lightgrey}\textbf{61.6} & \cellcolor{lightgrey}\textbf{48.9}\\
    \end{tabular}
    \end{center}}
    \end{minipage}
    }
\vspace{-3mm}
\caption{Comparison with the state-of-the-art CLIP-based methods on three datasets. ``\includegraphics[width=0.25cm,height=0.25cm]{fig/frozen.pdf}'': frozen backbone. \textbf{(a)} Zero-shot accuracy on HMDB51~\cite{jhuang2011hmdb51} and UCF101~\cite{soomro2012ucf101} across three splits.
\textbf{(b)} Results on the Epic-Kitchens-100~\cite{damen2020ek100} validation set.}
\end{table*}

\begin{table*}[t]
    \centering
    \tablestyle{4pt}{1.05}
\begin{tabular}{ccccccccc}
Method &\makecell{Pre-train} &  Architecture & Input Size &\makecell{TFLOPs$\times$Cr.$\times$Cl.} & \makecell{Param (M)}& Frozen &  \makecell{Top-1} & \makecell{Top-5}\\
\midrule[1.15pt]
SlowFast~\cite{feichtenhofer2019slowfast} & -  & R101+NL & $16\times224^2$ & $0.4\times3\times10$ & 60& \xmark  & 79.8 & 93.9\\
TimeSformer~\cite{bertasius2021timesformer} & ImageNet-21K  & ViT-L & $96\times224^2$ & $8.4\times3\times1$ & 430& \xmark  & 80.7 & 94.7 \\
MViT~\cite{fan2021mvit} & -  & MViT-B & $64\times224^2$ & $0.5\times1\times5$ & 37& \xmark  & 81.2 & 95.1 \\
ViViT FE~\cite{arnab2021vivit} & ImageNet-21K  & ViT-L & $128\times224^2$ & $4.0\times3\times1$& N/A& \xmark  & 81.7 & 93.8 \\
Video Swin~\cite{liu2022videoswin} & ImageNet-21K  & Swin-L & $32\times224^2$ & $0.6\times3\times4$ & 197& \xmark & 83.1 & 95.9 \\
TAdaConvNeXtV2~\cite{huang2023tadav2} & ImageNet-21K  & ConvNeXt-B & $32\times224^2$ & $0.3\times3\times4$ &  146& \xmark & 83.7 & - \\
\midrule
ST-Adapter\includegraphics[width=0.25cm,height=0.25cm]{fig/frozen.pdf}~\cite{pan2022st-adapter} & CLIP-400M  & ViT-B & $32\times224^2$ & $0.61\times1\times3$ & 93& \cmark & 82.7 & 96.2 \\
EVL\includegraphics[width=0.25cm,height=0.25cm]{fig/frozen.pdf}~\cite{lin2022evl} & CLIP-400M  & ViT-B & $32\times224^2$ & $0.59\times1\times3$  & 115& \cmark & 84.2 & - \\
\cellcolor{lightgrey}\textbf{DiST}$_{\gamma\text{=2}}$\includegraphics[width=0.25cm,height=0.25cm]{fig/frozen.pdf} &\cellcolor{lightgrey}CLIP-400M&\cellcolor{lightgrey}ViT-B &\cellcolor{lightgrey}$8\times224^2$ &\cellcolor{lightgrey}$0.16\times1\times3$ &\cellcolor{lightgrey}112&\cellcolor{lightgrey}\cmark&\cellcolor{lightgrey}83.6&\cellcolor{lightgrey}96.3 \\
\cellcolor{lightgrey}\textbf{DiST}$_{\gamma\text{=2}}$\includegraphics[width=0.25cm,height=0.25cm]{fig/frozen.pdf} &\cellcolor{lightgrey}CLIP-400M&\cellcolor{lightgrey}ViT-B &\cellcolor{lightgrey}$16\times224^2$ &\cellcolor{lightgrey}$0.32\times1\times3$ &\cellcolor{lightgrey}112&\cellcolor{lightgrey}\cmark&\cellcolor{lightgrey}84.4&\cellcolor{lightgrey}96.7 \\
\cellcolor{lightgrey}\textbf{DiST}$_{\gamma\text{=2}}$\includegraphics[width=0.25cm,height=0.25cm]{fig/frozen.pdf} &\cellcolor{lightgrey}CLIP-400M&\cellcolor{lightgrey}ViT-B &\cellcolor{lightgrey}$32\times224^2$ &\cellcolor{lightgrey}$0.65\times1\times3$ &\cellcolor{lightgrey}112&\cellcolor{lightgrey}\cmark&\cellcolor{lightgrey}85.0&\cellcolor{lightgrey}97.0 \\
\cellcolor{lightgrey}\textbf{DiST}$_{\gamma\text{=2}}$\includegraphics[width=0.25cm,height=0.25cm]{fig/frozen.pdf} &\cellcolor{lightgrey}\scriptsize CLIP-400M+K710&\cellcolor{lightgrey}ViT-B &\cellcolor{lightgrey}$32\times224^2$ &\cellcolor{lightgrey}$0.65\times1\times3$ &\cellcolor{lightgrey}112&\cellcolor{lightgrey}\cmark&\cellcolor{lightgrey}\textbf{86.8}&\cellcolor{lightgrey}\textbf{97.5} \\
\midrule
\textcolor{grey}{UnifromerV2}~\cite{li2022uniformerv2} & \textcolor{grey}{\scriptsize CLIP-400M+K710}  & \textcolor{grey}{ViT-L} & \textcolor{grey}{$32\times224^2$} & \textcolor{grey}{$2.66\times2\times3$}  & \textcolor{grey}{354}&\textcolor{grey}{\xmark} & \textcolor{grey}{89.3} & \textcolor{grey}{98.2} \\
\textcolor{grey}{TAdaFormer}~\cite{huang2023tadav2} & \textcolor{grey}{\scriptsize CLIP-400M+K710}  & \textcolor{grey}{ViT-L} & \textcolor{grey}{$32\times224^2$} & \textcolor{grey}{$1.41\times4\times3$}  & \textcolor{grey}{364}&\textcolor{grey}{\xmark} & \textcolor{grey}{89.5} & \textcolor{grey}{-} \\
ST-Adapter\includegraphics[width=0.25cm,height=0.25cm]{fig/frozen.pdf}~\cite{pan2022st-adapter} & CLIP-400M  & ViT-L & $32\times224^2$ & $2.75\times1\times3$& 347 & \cmark  & 87.2 & 97.6 \\
EVL\includegraphics[width=0.25cm,height=0.25cm]{fig/frozen.pdf}~\cite{lin2022evl} & CLIP-400M  & ViT-L & $32\times224^2$ & $2.70\times1\times3$ & 363& \cmark  & 87.3 & - \\
\cellcolor{lightgrey}\textbf{DiST}$_{\gamma\text{=2}}$\includegraphics[width=0.25cm,height=0.25cm]{fig/frozen.pdf} &\cellcolor{lightgrey}CLIP-400M&\cellcolor{lightgrey}ViT-L &\cellcolor{lightgrey}$8\times224^2$ &\cellcolor{lightgrey}$0.71\times1\times3$&\cellcolor{lightgrey}343& \cellcolor{lightgrey}\cmark&\cellcolor{lightgrey}86.9&\cellcolor{lightgrey}97.6\\
\cellcolor{lightgrey}\textbf{DiST}$_{\gamma\text{=2}}$\includegraphics[width=0.25cm,height=0.25cm]{fig/frozen.pdf} &\cellcolor{lightgrey}CLIP-400M&\cellcolor{lightgrey}ViT-L &\cellcolor{lightgrey}$16\times224^2$ &\cellcolor{lightgrey}$1.42\times1\times3$ &\cellcolor{lightgrey}343&\cellcolor{lightgrey}\cmark&\cellcolor{lightgrey}87.6&\cellcolor{lightgrey}97.8 \\
\cellcolor{lightgrey}\textbf{DiST}$_{\gamma\text{=2}}$\includegraphics[width=0.25cm,height=0.25cm]{fig/frozen.pdf} &\cellcolor{lightgrey}CLIP-400M&\cellcolor{lightgrey}ViT-L &\cellcolor{lightgrey}$32\times224^2$ &\cellcolor{lightgrey}$2.83\times1\times3$ &\cellcolor{lightgrey}343&\cellcolor{lightgrey}\cmark&\cellcolor{lightgrey}88.0&\cellcolor{lightgrey}97.9 \\
\cellcolor{lightgrey}\textbf{DiST}$_{\gamma\text{=2}}$\includegraphics[width=0.25cm,height=0.25cm]{fig/frozen.pdf} &\cellcolor{lightgrey}\scriptsize CLIP-400M+K710&\cellcolor{lightgrey}ViT-L &\cellcolor{lightgrey}$32\times224^2$ &\cellcolor{lightgrey}$2.83\times1\times3$&\cellcolor{lightgrey}343&\cellcolor{lightgrey}\cmark&\cellcolor{lightgrey}\textbf{89.5}&\cellcolor{lightgrey}\textbf{98.4} \\
\midrule
\textcolor{grey}{X-CLIP}~\cite{ni2022expanding-xclip} & \textcolor{grey}{CLIP-400M} & \textcolor{grey}{ViT-L} & \textcolor{grey}{$16\times336^2$} & \textcolor{grey}{$3.09\times3\times4$}& \textcolor{grey}{354}&\textcolor{grey}{\xmark}  & \textcolor{grey}{87.7} & \textcolor{grey}{97.4} \\
\textcolor{grey}{BIKE}~\cite{wu2023bidirectional} & \textcolor{grey}{CLIP-400M} & \textcolor{grey}{ViT-L} & \textcolor{grey}{$32\times336^2$} & \textcolor{grey}{$3.73\times3\times4$}& \textcolor{grey}{230}&\textcolor{grey}{\xmark}  & \textcolor{grey}{88.6} & \textcolor{grey}{98.3} \\
EVL\includegraphics[width=0.25cm,height=0.25cm]{fig/frozen.pdf}~\cite{lin2022evl} & CLIP-400M  & ViT-L & $32\times336^2$ & $6.07\times1\times3$ & 363 & \cmark & 87.7 & - \\
Text4Vis\includegraphics[width=0.25cm,height=0.25cm]{fig/frozen.pdf}~\cite{wu2023revisiting} & CLIP-400M  & ViT-L & $32\times336^2$ & $3.83\times1\times3$ & 231 & \cmark & 87.8 & 97.6 \\
UnifromerV2\includegraphics[width=0.25cm,height=0.25cm]{fig/frozen.pdf}~\cite{li2022uniformerv2} & \scriptsize CLIP-400M+K710  & ViT-L & $32\times336^2$ & $6.27\times1\times3$& 354&\cmark  & 88.8 & 98.1 \\
\cellcolor{lightgrey}\textbf{DiST}$_{\gamma\text{=2}}$\includegraphics[width=0.25cm,height=0.25cm]{fig/frozen.pdf} &\cellcolor{lightgrey}CLIP-400M&\cellcolor{lightgrey}ViT-L &\cellcolor{lightgrey}$32\times336^2$ &\cellcolor{lightgrey}$6.64\times1\times3$ &\cellcolor{lightgrey}343&\cellcolor{lightgrey}\cmark&\cellcolor{lightgrey}88.5&\cellcolor{lightgrey}98.2 \\
\cellcolor{lightgrey}\textbf{DiST}$_{\gamma\text{=2}}$\includegraphics[width=0.25cm,height=0.25cm]{fig/frozen.pdf} &\cellcolor{lightgrey}\scriptsize CLIP-400M+K710&\cellcolor{lightgrey}ViT-L &\cellcolor{lightgrey}$32\times336^2$ &\cellcolor{lightgrey}$6.64\times1\times3$&\cellcolor{lightgrey}343&\cellcolor{lightgrey}\cmark&\cellcolor{lightgrey}\textbf{89.7}&\cellcolor{lightgrey}\textbf{98.5} \\

\end{tabular}
    \caption{Comparison with state-of-the-arts on Kinetics-400. }
    \label{tab:sota_k400}
\end{table*}

\noindent
\textbf{Has the temporal encoder learned video-specific representations?}
To demonstrate this, we first utilized CKA similarity~\cite{cka} to analyze the feature correlation between various video features and the CLIP pre-trained image features.
As shown in Fig.~\ref{fig:cka_vis_param}~(a), we can see that the video features learned by EVL~\cite{lin2022evl} are highly correlated with the image features generated by CLIP. 
This explains the reason why EVL has weak temporal modeling ability.
However, the correlations of our integrated feature, temporal feature are gradually weakened, which fully demonstrates that decoupling spatio-temporal learning indeed enables the temporal encoder to capture  temporal patterns largely complementary to spatial features.
In Fig.\ref{fig:cka_vis_param}~(b), we further visualize the activation amplitude of the features from the temporal encoder and the spatial encoder. 
As can be seen, the temporal is more sensitive to the inter-frame motion, thus facilitating the learning of video-specific features.

\noindent
\textbf{Optional designs for feature interactions.} 
In Tab.~\ref{tab:optional_design}, we evaluate the downsampling ways in function $\Psi(\cdot)$ and upsampling ways in function $\Phi(\cdot)$ for feature interactions. 
First, for downsampling, the learnable temporal convolution (\ie, DConv) slightly outperformed the pooling methods by around 0.3\%. 
Intriguingly, for the upsampling methods, the performance with the learnable deconvolution is worse. We speculate that the deconvolution and trilinear incorporates adjacent frames, resulting in spatial semantic shifts. We thus employ DConv and nearest as our default.


\noindent
\textbf{Training and inference consumption.} In Tab.~\ref{tab:train_infer_compare}, we compare the training and inference time with existing efficient fine-tuning approaches under the same hardware. 
Firstly, in training, as a back-propagation-free approach, the GPU memory consumption of our DiST is merely 32\% (\ie, 12.68G \textit{v.s.} 39.10G) of ST-Adapter~\cite{pan2022st-adapter}. 
Benefiting from the lightweight design of the temporal encoder, the training step time is only 73\% of EVL~\cite{lin2022evl}, which is also based on the back-propagation-free backbone.
Moreover, the training epochs required by DiST is also less than that of EVL~\cite{lin2022evl} and ST-Adapter~\cite{pan2022st-adapter}, since the complete reliance on image-specific features for spatio-temporal learning may potentially pose additional challenges.
Secondly, in inference, we test the throughput for the above methods with batch size of 32. 
Since ST-Adapter~\cite{pan2022st-adapter} has no additional branches, its throughput is slightly higher than our DiST. 
However, compared with the similar EVL, our throughput is increased by 1.26 $\times$ (\ie, from 38 Videos/s to 48 Videos/s).

\subsection{Comparison with State-of-the-art}

\noindent
\textbf{Zero-shot experiments.} 
Due to the retention of the frozen text branch, our approach is still able to conduct zero-shot tasks. 
We employ the 32-frame Kinetics-400 fine-tuned models in Tab.~\ref{tab:sota_k400} for evaluation.
As in Tab.~\ref{tab:zero_shot}, with the same pre-trained model (\ie, ViT-B/16), our method remarkably outperforms existing fully fine-tuned X-CLIP~\cite{ni2022expanding-xclip} by 10.8\% on HMDB51 and is more stable across different splits.
The relatively minor improvement on UCF101 is attributed to the spatially-focused dataset with limited temporal clues available for utilization. 
Furthermore, DiST with larger models can achieve consistent performance gains which can be attributed to the excellent architectural scalability of DiST.

\noindent
\textbf{Egocentric action recognition.} Tab.~\ref{tab:ek100} presents fair comparisons between our DiST and existing Frozen-CLIP approaches. 
It is evident that DiST consistently demonstrates a convincing performance advantage.
With the same ViT-B/16 as spatial encoder, DiST achieves accuracy improvements of over ST-Adapter~\cite{pan2022st-adapter} by 1.9\% and 3.1\% on verbs and nouns, respectively. 
This is attributed to decoupling temporal encoder to learn representations that complement frozen spatial features. 
Moreover, DiST continues to provide sustained benefits even on larger models.

\noindent
\textbf{Video recognition on SSV2 and K400.} 
First, for the temporally-heavy dataset, \ie, SSV2 in Tab.~\ref{tab:sota_ssv2}, DiST outperforms other CLIP-based efficient fine-tuning approach by a notable margin. For example, compared with EVL~\cite{lin2022evl} that also uses the frozen CLIP features, DiST surpasses it by 8.5\% with a 32-frame ViT-B.
Compared with fully fine-tuned UniformerV2~\cite{li2022uniformerv2}, our efficient DiST still achieve comparable accuracy.
Second, on the spatially-heavy K400\cite{kay2017k400}, DiST is still highly competitive. Compared with EVL with better performance, DiST can always achieve improvements around 0.8\% regardless of the pre-training models. 
With these observations, we can summarize that DiST enjoys the dual advantages of spatial modeling and temporal modeling.
Besides, following UniformerV2, we pre-train the lightweight temporal encoder and integration branch on a large-scale video dataset, \ie, Kinetics-710~\cite{kay2017k400,carreira2018k600,carreira2019k700}, and the performances are further improved. 
%
When inputting 32 frames with $336 \times 336$ size, our approach exceeds UniformerV2 by 0.9\% using ViT-L model, which can demonstrate the strong data scalability of DiST.

\section{Conclusion}

In this work, we propose DiST, an image-to-video transfer learning framework that enjoys both training efficiency and powerful temporal modeling capabilities. 
It is a dual-encoder structure, which includes a frozen but heavy spatial encoder and a lightweight learnable temporal encoder. 
Then, an integration branch fuses the spatial and temporal information into the unified spatio-temporal representations for video understanding.
Extensive experiments verify the scalability of DiST in both model size and data scale. We hope that our DiST can provide some inspiration for researchers who are interested in large-scale video models.

\section{Acknowledgement}

This work is supported by the National Natural Science Foundation of China under grant U22B2053 and 62176097, and by Alibaba DAMO Academy through Alibaba Research Intern Program.

{\small
\bibliographystyle{ieee_fullname}
\bibliography{egbib}
}

\newpage
\appendix
\renewcommand{\thetable}{A\arabic{table}}
\renewcommand{\thefigure}{A\arabic{figure}}
\renewcommand{\theequation}{A\arabic{equation}}
\include{appendix}

\end{document}

%% file: appendix.tex
\section*{Overview}
In this supplementary material, we first provide more ablation studies, and comparison with state-of-the-art approaches on zero-shot task and video recognition task in Sec.~\ref{sec:add_results}. Then, the implementation details for SSV2, K400 and EK100 are presented in Sec.~\ref{sec:impl_details}.


\section{Additional Results}
\label{sec:add_results}
\subsection{Ablation Studies}

\begin{table}[h!]
    \centering
    \tablestyle{4pt}{1.05}
    \begin{tabular}{ccc|cc}
        Method &    Pretraining & Frozen & SSV2 & K400 \\
        \shline
        EVL~\cite{lin2022evl} & ImageNet-21k & \cmark & N/A &  75.4 \\
        ST-Adapter~\cite{pan2022st-adapter} & ImageNet-21k& \cmark & 62.8&  76.6 \\
        TimeSformer~\cite{bertasius2021timesformer}& ImageNet-21k & \xmark & 59.5 &  78.0 \\
        X-ViT~\cite{bulat2021xvit}& ImageNet-21k & \xmark & 64.4 &  78.5 \\
        \cellcolor{lightgrey}\textbf{DiST} & \cellcolor{lightgrey}ImageNet-21k& \cellcolor{lightgrey}\cmark & \cellcolor{lightgrey}\textbf{66.8} & \cellcolor{lightgrey}\textbf{79.8} \\
        \hline
        EVL~\cite{lin2022evl} & CLIP& \cmark & 61.0 &82.9  \\
        ST-Adapter~\cite{pan2022st-adapter} & CLIP& \cmark & 66.3& 82.0 \\
        \cellcolor{lightgrey}\textbf{DiST} & \cellcolor{lightgrey}CLIP& \cellcolor{lightgrey}\cmark & \cellcolor{lightgrey}\textbf{68.7} & \cellcolor{lightgrey}\textbf{83.6} \\
    \end{tabular}
    \caption{Comparsion with state-of-the-art under different pre-trained image encoders.}
    \label{tab:in21k_pretraining}
\end{table}


\noindent
\textbf{Fine-tuning with ImageNet pretrained models.} 
In fact, our DiST is not limited to CLIP pre-trained image models, and thus we have attempted to explore fine-tuning video models based on ImageNet supervised pre-training by following existing methods~\cite{lin2022evl,pan2022st-adapter}.
As shown in Tab.~\ref{tab:in21k_pretraining}, DiST outperforms the similar frozen CLIP-based fine-tuning method, \ie, EVL~\cite{lin2022evl} by 4.4 \%.
Compared to the adapter-based approach, \ie, ST-Adapter~\cite{pan2022st-adapter}, we exceed by 4.0\% and 2.2\% on SSV2 and K400 datasets, respectively. 
Besides, DiST also shows performance advantages over full fine-tuning methods, such as TimeSformer~\cite{bertasius2021timesformer} and X-ViT~\cite{bulat2021xvit}.
These results indicate that DiST is a more general network for image-to-video transfer learning.

\begin{table}[h!]
    \centering
    \begin{tabular}{cccc|cc}
        Frame  & 1-4 & 5-8 & 9-12  &  SSV2 & K400\\
        \shline
        \xmark & \xmark & \xmark & \cmark  & 66.0 & 83.0 \\
        \xmark & \xmark & \cmark & \cmark  & 67.8 & 83.3 \\
        \xmark & \cmark & \cmark & \cmark  & 68.0 & 83.4\\
        \cellcolor{lightgrey}\cmark & \cellcolor{lightgrey}\cmark & \cellcolor{lightgrey}\cmark & \cellcolor{lightgrey}\cmark  & \cellcolor{lightgrey}\textbf{68.7} & \cellcolor{lightgrey}\textbf{83.6}\\
    \end{tabular}
    \caption{Different layer of features.}
    \label{tab:different_feat}
\end{table}

\noindent
\textbf{Different depth features.}
%
As shown in Tab.~\ref{tab:different_feat}, if only using the deep features (\ie, 9-12) of ViT, its performance on the temporally heavy dataset (\ie, SSV2) is weak.
Nevertheless, the introduction of low-level features can bring up to 2.8\% gains at most.
This implies that the rich temporal details in the lower-level features are more beneficial for spatio-temporal learning. 

\begin{figure}[h!]
    \centering
    \includegraphics[width=1.0\linewidth]{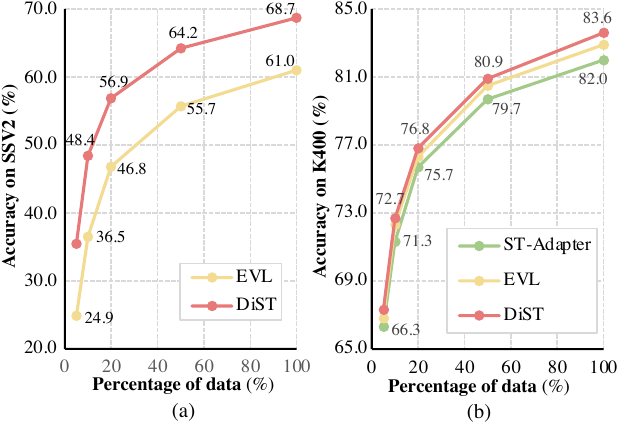}
    \caption{Performance comparison on varying training data scales.}
    \label{fig:data_efficiency}
\end{figure}

\begin{table*}[h!]
    \centering
    \tablestyle{4pt}{1.05}
\begin{tabular}{ccccccccc}
Method &\makecell{Pre-train} &  Architecture & Input Size & \makecell{FLOPs$\times$Cr.$\times$Cl. (T)} &  \makecell{Param (M)}& Frozen&  \makecell{Top-1} & \makecell{Top-5}\\
\midrule[1.15pt]
SlowFast~\cite{feichtenhofer2019slowfast} & ImageNet-21K  & R101+NL & $16\times224^2$ & $0.1\times3\times1$ & 60& \xmark  & 63.1 & 87.6 \\
ViViT FE~\cite{arnab2021vivit} & IN21K+K400  & ViT-L & $16\times224^2$ & $1.0\times3\times4$& 612& \xmark  & 65.4 & 89.8 \\
TAdaConvNeXt-T~\cite{huang2021tada} & ImageNet-1K  & ConvNeXt-T & $32\times224^2$ & $0.1\times3\times2$ &  38& \xmark & 67.1 & 90.4 \\
MTV-B(320p)~\cite{yan2022multiview}& IN21K+K400 & - &$32\times224^2$ & $0.9\times3\times4$& 310& \xmark & 68.5 & 90.4 \\
MViT~\cite{fan2021mvit} & Kinetics-600  & MViT-B-24 & $32\times224^2$ & $0.2\times3\times1$ & 53& \xmark  & 68.7 & 91.5 \\
Video Swin~\cite{liu2022videoswin} & IN21K+K400  & Swin-B & $32\times224^2$ & $0.3\times3\times1$ & 60& \xmark & 69.6 & 92.7 \\
\midrule
\textcolor{grey}{UnifromerV2}~\cite{li2022uniformerv2} & \textcolor{grey}{CLIP-400M}  & \textcolor{grey}{ViT-B} & \textcolor{grey}{$32\times224^2$} & \textcolor{grey}{$0.37\times1\times3$}  & \textcolor{grey}{163}&\textcolor{grey}{\xmark} & \textcolor{grey}{70.7 }& \textcolor{grey}{93.2} \\
EVL\includegraphics[width=0.25cm,height=0.25cm]{fig/frozen.pdf}~\cite{lin2022evl} & CLIP-400M  & ViT-B & $32\times224^2$ & $0.68\times1\times3$  & 175& \cmark & 62.4 & - \\
ST-Adapter\includegraphics[width=0.25cm,height=0.25cm]{fig/frozen.pdf}~\cite{pan2022st-adapter} & CLIP-400M  & ViT-B & $32\times224^2$ & $0.61\times1\times3$ & 93& \cmark & 69.5 & \textbf{92.6} \\
\cellcolor{lightgrey}\textbf{DiST}$_{\gamma\text{=2}}$\includegraphics[width=0.25cm,height=0.25cm]{fig/frozen.pdf} &\cellcolor{lightgrey}CLIP-400M&\cellcolor{lightgrey}ViT-B &\cellcolor{lightgrey}$8\times224^2$ &\cellcolor{lightgrey}$0.16\times1\times3$ &\cellcolor{lightgrey}105&\cellcolor{lightgrey}\cmark&\cellcolor{lightgrey}68.7&\cellcolor{lightgrey}91.1\\
\cellcolor{lightgrey}\textbf{DiST}$_{\gamma\text{=2}}$\includegraphics[width=0.25cm,height=0.25cm]{fig/frozen.pdf} &\cellcolor{lightgrey}CLIP-400M&\cellcolor{lightgrey}ViT-B &\cellcolor{lightgrey}$16\times224^2$ &\cellcolor{lightgrey}$0.32\times1\times3$ &\cellcolor{lightgrey}105&\cellcolor{lightgrey}\cmark&\cellcolor{lightgrey}70.2&\cellcolor{lightgrey}92.0\\
\cellcolor{lightgrey}\textbf{DiST}$_{\gamma\text{=2}}$\includegraphics[width=0.25cm,height=0.25cm]{fig/frozen.pdf} &\cellcolor{lightgrey}CLIP-400M&\cellcolor{lightgrey}ViT-B &\cellcolor{lightgrey}$32\times224^2$ &\cellcolor{lightgrey}$0.65\times1\times3$ &\cellcolor{lightgrey}105&\cellcolor{lightgrey}\cmark&\cellcolor{lightgrey}\textbf{70.9}&\cellcolor{lightgrey}92.1\\
\midrule
\textcolor{grey}{UnifromerV2}~\cite{li2022uniformerv2} & \textcolor{grey}{CLIP-400M}  & \textcolor{grey}{ViT-L} & \textcolor{grey}{$32\times224^2$} & \textcolor{grey}{$1.73\times1\times3$}  & \textcolor{grey}{574}&\textcolor{grey}{\xmark} & \textcolor{grey}{73.0} & \textcolor{grey}{94.5} \\
\textcolor{grey}{TAdaFormer}~\cite{huang2023tadav2} & \textcolor{grey}{CLIP-400M}  & \textcolor{grey}{ViT-L} & \textcolor{grey}{$32\times224^2$} & \textcolor{grey}{$1.70\times2\times3$}  & \textcolor{grey}{364}&\textcolor{grey}{\xmark} & \textcolor{grey}{73.6} & \textcolor{grey}{-} \\
EVL\includegraphics[width=0.25cm,height=0.25cm]{fig/frozen.pdf}~\cite{lin2022evl} & CLIP-400M  & ViT-L & $32\times224^2$ & $3.21\times1\times3$ & 654& \cmark  & 66.7 &  -\\
ST-Adapter\includegraphics[width=0.25cm,height=0.25cm]{fig/frozen.pdf}~\cite{pan2022st-adapter} & CLIP-400M  & ViT-L & $32\times224^2$ & $2.75\times1\times3$& 347 & \cmark  & 72.3 & \textbf{93.9} \\
\cellcolor{lightgrey}\textbf{DiST}$_{\gamma\text{=2}}$\includegraphics[width=0.25cm,height=0.25cm]{fig/frozen.pdf} &\cellcolor{lightgrey}CLIP-400M&\cellcolor{lightgrey}ViT-L &\cellcolor{lightgrey}$8\times224^2$ &\cellcolor{lightgrey}$0.71\times1\times3$&\cellcolor{lightgrey}336& \cellcolor{lightgrey}\cmark&\cellcolor{lightgrey}70.8&\cellcolor{lightgrey}92.3\\
\cellcolor{lightgrey}\textbf{DiST}$_{\gamma\text{=2}}$\includegraphics[width=0.25cm,height=0.25cm]{fig/frozen.pdf} &\cellcolor{lightgrey}CLIP-400M&\cellcolor{lightgrey}ViT-L &\cellcolor{lightgrey}$16\times224^2$ &\cellcolor{lightgrey}$1.42\times1\times3$ &\cellcolor{lightgrey}336&\cellcolor{lightgrey}\cmark&\cellcolor{lightgrey}72.5&\cellcolor{lightgrey}93.0\\
\cellcolor{lightgrey}\textbf{DiST}$_{\gamma\text{=2}}$\includegraphics[width=0.25cm,height=0.25cm]{fig/frozen.pdf} &\cellcolor{lightgrey}CLIP-400M&\cellcolor{lightgrey}ViT-L &\cellcolor{lightgrey}$32\times224^2$ &\cellcolor{lightgrey}$2.83\times1\times3$ &\cellcolor{lightgrey}336&\cellcolor{lightgrey}\cmark&\cellcolor{lightgrey}\textbf{73.1}&\cellcolor{lightgrey}93.2\\
\midrule
EVL\includegraphics[width=0.25cm,height=0.25cm]{fig/frozen.pdf}~\cite{lin2022evl} & CLIP-400M  & ViT-L & $32\times336^2$ & $8.08\times1\times3$ & 654 & \cmark & 68.0 & - \\
\cellcolor{lightgrey}\textbf{DiST}$_{\gamma\text{=2}}$\includegraphics[width=0.25cm,height=0.25cm]{fig/frozen.pdf} &\cellcolor{lightgrey}CLIP-400M&\cellcolor{lightgrey}ViT-L &\cellcolor{lightgrey}$8\times336^2$ &\cellcolor{lightgrey}$1.66\times1\times3$ &\cellcolor{lightgrey}336&\cellcolor{lightgrey}\cmark&\cellcolor{lightgrey}{71.2}&\cellcolor{lightgrey}{92.5}\\
\cellcolor{lightgrey}\textbf{DiST}$_{\gamma\text{=2}}$\includegraphics[width=0.25cm,height=0.25cm]{fig/frozen.pdf} &\cellcolor{lightgrey}CLIP-400M&\cellcolor{lightgrey}ViT-L &\cellcolor{lightgrey}$16\times336^2$ &\cellcolor{lightgrey}$3.32\times1\times3$ &\cellcolor{lightgrey}336&\cellcolor{lightgrey}\cmark&\cellcolor{lightgrey}{72.6}&\cellcolor{lightgrey}{93.0}\\
\cellcolor{lightgrey}\textbf{DiST}$_{\gamma\text{=2}}$\includegraphics[width=0.25cm,height=0.25cm]{fig/frozen.pdf} &\cellcolor{lightgrey}CLIP-400M&\cellcolor{lightgrey}ViT-L &\cellcolor{lightgrey}$32\times336^2$ &\cellcolor{lightgrey}$6.64\times1\times3$ &\cellcolor{lightgrey}336&\cellcolor{lightgrey}\cmark&\cellcolor{lightgrey}\textbf{73.3}&\cellcolor{lightgrey}\textbf{93.5}\\

\end{tabular}
    \caption{Comparison with the state-of-the-art methods on Something-Something V2. ``Cr.'' and ``Cl.'' are the abbreviation for ``spatial crops'' and ``temporal clips''. ``Frozen'' indicates freezing the CLIP pre-trained parameters.}
    \label{tab:sota_ssv2_appendix}
\end{table*}

\noindent
\textbf{Data efficiency.} 
Data efficiency refers to the utilization efficiency of limited data by fine-tuning the models with only a portion of the training data.
In Fig.~\ref{fig:data_efficiency}, DiST and other clip-based pre-training methods~\cite{pan2022st-adapter,lin2022evl} are compared in terms of data efficiency on SSV2 and K400 datasets.
As can be seen, our DiST exhibits consistent advantages over existing ST-Adapter~\cite{pan2022st-adapter} and EVL~\cite{lin2022evl} across different proportions of training data. 
Especially on the temporally dependent dataset, \ie, SSV2, DiST demonstrates improvements of approximately 10\% over EVL~\cite{lin2022evl} with fewer training scales.
This suggests that DiST is easier to fine-tune and shows better generalization capability.

\begin{table}[h!]
    \centering
    \small    
    \tablestyle{4pt}{1.}
    \begin{tabular}{ccc|cc}
        Method  & Model  & Frames &  HMDB51 & UCF101\\
        \shline
       ActionCLIP~\cite{wang2021actionclip} & B/16 & 32$\times$1$\times$1 & 40.8$\pm$5.4 &58.3$\pm$3.4 \\
       X-CLIP~\cite{ni2022expanding-xclip} & B/16 &32$\times$1$\times$1 & 44.6$\pm$5.2 &72.0$\pm$2.3 \\
       \cellcolor{lightgrey}\textbf{DiST}$_{\gamma\text{=2}}$\includegraphics[width=0.25cm,height=0.25cm]{fig/frozen.pdf} &  \cellcolor{lightgrey}B/16&  32$\times$1$\times$1\cellcolor{lightgrey}&\cellcolor{lightgrey}55.4$\pm$1.2 & \cellcolor{lightgrey}72.3$\pm$0.6 \\
       \cellcolor{lightgrey}$^{\text{\textbf{\dag}}}$\textbf{DiST}$_{\gamma\text{=2}}$\includegraphics[width=0.25cm,height=0.25cm]{fig/frozen.pdf} &  \cellcolor{lightgrey}B/16& 32$\times$1$\times$1\cellcolor{lightgrey}&\cellcolor{lightgrey}57.4$\pm$0.9 & \cellcolor{lightgrey}73.2$\pm$0.6 \\
       \cellcolor{lightgrey}\textbf{DiST}$_{\gamma\text{=2}}$\includegraphics[width=0.25cm,height=0.25cm]{fig/frozen.pdf} &  \cellcolor{lightgrey}L/14& \cellcolor{lightgrey}32$\times$1$\times$1&\cellcolor{lightgrey}{57.5}$\pm$1.6&\cellcolor{lightgrey}{74.9}$\pm$0.8 \\
       \cellcolor{lightgrey}$^{\text{\textbf{\dag}}}$\textbf{DiST}$_{\gamma\text{=2}}$\includegraphics[width=0.25cm,height=0.25cm]{fig/frozen.pdf} &  \cellcolor{lightgrey}L/14  &\cellcolor{lightgrey}32$\times$1$\times$1&\cellcolor{lightgrey}\textbf{61.8}$\pm$1.3&\cellcolor{lightgrey}\textbf{75.8}$\pm$0.7 \\

    \end{tabular}
    \caption{Zero-shot performance on HMDB51~\cite{jhuang2011hmdb51} and UCF101~\cite{soomro2012ucf101} across three splits. $^{\text{\textbf{\dag}}}$ indicates Kinetics-710~\cite{kay2017k400,carreira2018k600,carreira2019k700} pre-trained models.}
    \label{tab:zero_shot_appendix}
\end{table}

\subsection{Comparison with the state-of-the-art methods}

\noindent
\textbf{Zero-shot accuracy with Kinetics-710 pre-training.} 
We further evaluate the zero-shot performance of DiST on HMDB51~\cite{jhuang2011hmdb51} and UCF101~\cite{soomro2012ucf101} with the large-scale video dataset, \ie, Kinetics-710~\cite{kay2017k400,carreira2018k600,carreira2019k700}, pre-trained models.
As shown in Tab.~\ref{tab:zero_shot_appendix}, we can observe that regardless of the model size, remarkable improvements can be achieved by fine-tuning our lightweight temporal encoder and integration branch on Kinetics-710. 
Particularly, on HMDB51 that relies on temporal information, the gains of ViT-B and ViT-L can reach 2.0\% and 4.3\%, respectively. 
This further demonstrates the scalability of DiST on both data scale and model size.

\begin{table*}[h!]
    \centering
    \tablestyle{4pt}{1.05}
\begin{tabular}{ccccccccc}
Method &\makecell{Pre-train} &  Architecture & Input Size &\makecell{TFLOPs$\times$Cr.$\times$Cl.} & \makecell{Param (M)}& Frozen &  \makecell{Top-1} & \makecell{Top-5}\\
\midrule[1.15pt]
TAda~\cite{huang2021tada} & ImageNet-1K  & ConvNeXt-T & $32\times224^2$ & $0.1\times3\times2$ &  38& \xmark & 79.1 & 93.7\\
SlowFast~\cite{feichtenhofer2019slowfast} & -  & R101+NL & $16\times224^2$ & $0.4\times3\times10$ & 60& \xmark  & 79.8 & 93.9\\
TimeSformer~\cite{bertasius2021timesformer} & ImageNet-21K  & ViT-L & $96\times224^2$ & $8.4\times3\times1$ & 430& \xmark  & 80.7 & 94.7 \\
MViT~\cite{fan2021mvit} & -  & MViT-B & $64\times224^2$ & $0.5\times1\times5$ & 37& \xmark  & 81.2 & 95.1 \\
ViViT FE~\cite{arnab2021vivit} & ImageNet-21K  & ViT-L & $128\times224^2$ & $4.0\times3\times1$& N/A& \xmark  & 81.7 & 93.8 \\
Video Swin~\cite{liu2022videoswin} & ImageNet-21K  & Swin-L & $32\times224^2$ & $0.6\times3\times4$ & 197& \xmark & 83.1 & 95.9 \\
\midrule
\textcolor{grey}{UnifromerV2}~\cite{li2022uniformerv2} & \scriptsize \textcolor{grey}{CLIP-400M+K710}  & \textcolor{grey}{ViT-B} & \textcolor{grey}{$8\times224^2$} & \textcolor{grey}{$0.13\times1\times3$}  & \textcolor{grey}{115}&\textcolor{grey}{\xmark} & \textcolor{grey}{85.2} & \textcolor{grey}{96.7} \\
ST-Adapter\includegraphics[width=0.25cm,height=0.25cm]{fig/frozen.pdf}~\cite{pan2022st-adapter} & CLIP-400M  & ViT-B & $32\times224^2$ & $0.61\times1\times3$ & 93& \cmark & 82.7 & 96.2 \\
EVL\includegraphics[width=0.25cm,height=0.25cm]{fig/frozen.pdf}~\cite{lin2022evl} & CLIP-400M  & ViT-B & $32\times224^2$ & $0.59\times1\times3$  & 115& \cmark & 84.2 & - \\
\cellcolor{lightgrey}\textbf{DiST}$_{\gamma\text{=2}}$\includegraphics[width=0.25cm,height=0.25cm]{fig/frozen.pdf} &\cellcolor{lightgrey}CLIP-400M&\cellcolor{lightgrey}ViT-B &\cellcolor{lightgrey}$8\times224^2$ &\cellcolor{lightgrey}$0.16\times1\times3$ &\cellcolor{lightgrey}112&\cellcolor{lightgrey}\cmark&\cellcolor{lightgrey}83.6&\cellcolor{lightgrey}96.3 \\
\cellcolor{lightgrey}\textbf{DiST}$_{\gamma\text{=2}}$\includegraphics[width=0.25cm,height=0.25cm]{fig/frozen.pdf} &\cellcolor{lightgrey}CLIP-400M&\cellcolor{lightgrey}ViT-B &\cellcolor{lightgrey}$16\times224^2$ &\cellcolor{lightgrey}$0.32\times1\times3$ &\cellcolor{lightgrey}112&\cellcolor{lightgrey}\cmark&\cellcolor{lightgrey}84.4&\cellcolor{lightgrey}96.7 \\
\cellcolor{lightgrey}\textbf{DiST}$_{\gamma\text{=2}}$\includegraphics[width=0.25cm,height=0.25cm]{fig/frozen.pdf} &\cellcolor{lightgrey}CLIP-400M&\cellcolor{lightgrey}ViT-B &\cellcolor{lightgrey}$32\times224^2$ &\cellcolor{lightgrey}$0.65\times1\times3$ &\cellcolor{lightgrey}112&\cellcolor{lightgrey}\cmark&\cellcolor{lightgrey}85.0&\cellcolor{lightgrey}97.0 \\
\cellcolor{lightgrey}\textbf{DiST}$_{\gamma\text{=2}}$\includegraphics[width=0.25cm,height=0.25cm]{fig/frozen.pdf} &\cellcolor{lightgrey}\scriptsize CLIP-400M+K710&\cellcolor{lightgrey}ViT-B &\cellcolor{lightgrey}$8\times224^2$ &\cellcolor{lightgrey}$0.16\times1\times3$ &\cellcolor{lightgrey}112&\cellcolor{lightgrey}\cmark&\cellcolor{lightgrey}85.1&\cellcolor{lightgrey}96.8 \\
\cellcolor{lightgrey}\textbf{DiST}$_{\gamma\text{=2}}$\includegraphics[width=0.25cm,height=0.25cm]{fig/frozen.pdf} &\cellcolor{lightgrey}\scriptsize CLIP-400M+K710&\cellcolor{lightgrey}ViT-B &\cellcolor{lightgrey}$16\times224^2$ &\cellcolor{lightgrey}$0.32\times1\times3$ &\cellcolor{lightgrey}112&\cellcolor{lightgrey}\cmark&\cellcolor{lightgrey}85.8&\cellcolor{lightgrey}97.2 \\
\cellcolor{lightgrey}\textbf{DiST}$_{\gamma\text{=2}}$\includegraphics[width=0.25cm,height=0.25cm]{fig/frozen.pdf} &\cellcolor{lightgrey}\scriptsize CLIP-400M+K710&\cellcolor{lightgrey}ViT-B &\cellcolor{lightgrey}$32\times224^2$ &\cellcolor{lightgrey}$0.65\times1\times3$ &\cellcolor{lightgrey}112&\cellcolor{lightgrey}\cmark&\cellcolor{lightgrey}\textbf{86.8}&\cellcolor{lightgrey}\textbf{97.5} \\
\midrule
\textcolor{grey}{UnifromerV2}~\cite{li2022uniformerv2} & \textcolor{grey}{\scriptsize CLIP-400M+K710}  & \textcolor{grey}{ViT-L} & \textcolor{grey}{$32\times224^2$} & \textcolor{grey}{$2.66\times2\times3$}  & \textcolor{grey}{354}&\textcolor{grey}{\xmark} & \textcolor{grey}{89.3} & \textcolor{grey}{98.2} \\
\textcolor{grey}{TAdaFormer}~\cite{huang2023tadav2} & \textcolor{grey}{\scriptsize CLIP-400M+K710}  & \textcolor{grey}{ViT-L} & \textcolor{grey}{$32\times224^2$} & \textcolor{grey}{$1.41\times4\times3$}  & \textcolor{grey}{364}&\textcolor{grey}{\xmark} & \textcolor{grey}{89.5} & \textcolor{grey}{-} \\
ST-Adapter\includegraphics[width=0.25cm,height=0.25cm]{fig/frozen.pdf}~\cite{pan2022st-adapter} & CLIP-400M  & ViT-L & $32\times224^2$ & $2.75\times1\times3$& 347 & \cmark  & 87.2 & 97.6 \\
EVL\includegraphics[width=0.25cm,height=0.25cm]{fig/frozen.pdf}~\cite{lin2022evl} & CLIP-400M  & ViT-L & $32\times224^2$ & $2.70\times1\times3$ & 363& \cmark  & 87.3 & - \\
\cellcolor{lightgrey}\textbf{DiST}$_{\gamma\text{=2}}$\includegraphics[width=0.25cm,height=0.25cm]{fig/frozen.pdf} &\cellcolor{lightgrey}CLIP-400M&\cellcolor{lightgrey}ViT-L &\cellcolor{lightgrey}$8\times224^2$ &\cellcolor{lightgrey}$0.71\times1\times3$&\cellcolor{lightgrey}343& \cellcolor{lightgrey}\cmark&\cellcolor{lightgrey}86.9&\cellcolor{lightgrey}97.6\\
\cellcolor{lightgrey}\textbf{DiST}$_{\gamma\text{=2}}$\includegraphics[width=0.25cm,height=0.25cm]{fig/frozen.pdf} &\cellcolor{lightgrey}CLIP-400M&\cellcolor{lightgrey}ViT-L &\cellcolor{lightgrey}$16\times224^2$ &\cellcolor{lightgrey}$1.42\times1\times3$ &\cellcolor{lightgrey}343&\cellcolor{lightgrey}\cmark&\cellcolor{lightgrey}87.6&\cellcolor{lightgrey}97.8 \\
\cellcolor{lightgrey}\textbf{DiST}$_{\gamma\text{=2}}$\includegraphics[width=0.25cm,height=0.25cm]{fig/frozen.pdf} &\cellcolor{lightgrey}CLIP-400M&\cellcolor{lightgrey}ViT-L &\cellcolor{lightgrey}$32\times224^2$ &\cellcolor{lightgrey}$2.83\times1\times3$ &\cellcolor{lightgrey}343&\cellcolor{lightgrey}\cmark&\cellcolor{lightgrey}88.0&\cellcolor{lightgrey}97.9 \\
\cellcolor{lightgrey}\textbf{DiST}$_{\gamma\text{=2}}$\includegraphics[width=0.25cm,height=0.25cm]{fig/frozen.pdf} &\cellcolor{lightgrey}\scriptsize CLIP-400M+K710&\cellcolor{lightgrey}ViT-L &\cellcolor{lightgrey}$8\times224^2$ &\cellcolor{lightgrey}$0.71\times1\times3$&\cellcolor{lightgrey}343&\cellcolor{lightgrey}\cmark&\cellcolor{lightgrey}87.8&\cellcolor{lightgrey}97.9 \\
\cellcolor{lightgrey}\textbf{DiST}$_{\gamma\text{=2}}$\includegraphics[width=0.25cm,height=0.25cm]{fig/frozen.pdf} &\cellcolor{lightgrey}\scriptsize CLIP-400M+K710&\cellcolor{lightgrey}ViT-L &\cellcolor{lightgrey}$16\times224^2$ &\cellcolor{lightgrey}$1.42\times1\times3$&\cellcolor{lightgrey}343&\cellcolor{lightgrey}\cmark&\cellcolor{lightgrey}88.6&\cellcolor{lightgrey}98.2 \\
\cellcolor{lightgrey}\textbf{DiST}$_{\gamma\text{=2}}$\includegraphics[width=0.25cm,height=0.25cm]{fig/frozen.pdf} &\cellcolor{lightgrey}\scriptsize CLIP-400M+K710&\cellcolor{lightgrey}ViT-L &\cellcolor{lightgrey}$32\times224^2$ &\cellcolor{lightgrey}$2.83\times1\times3$&\cellcolor{lightgrey}343&\cellcolor{lightgrey}\cmark&\cellcolor{lightgrey}\textbf{89.5}&\cellcolor{lightgrey}\textbf{98.4} \\
\midrule
\textcolor{grey}{X-CLIP}~\cite{ni2022expanding-xclip} & \textcolor{grey}{CLIP-400M} & \textcolor{grey}{ViT-L} & \textcolor{grey}{$16\times336^2$} & \textcolor{grey}{$3.09\times3\times4$}& \textcolor{grey}{354}&\textcolor{grey}{\xmark}  & \textcolor{grey}{87.7} & \textcolor{grey}{97.4} \\
\textcolor{grey}{BIKE}~\cite{wu2023bidirectional} & \textcolor{grey}{CLIP-400M} & \textcolor{grey}{ViT-L} & \textcolor{grey}{$32\times336^2$} & \textcolor{grey}{$3.73\times3\times4$}& \textcolor{grey}{230}&\textcolor{grey}{\xmark}  & \textcolor{grey}{88.6} & \textcolor{grey}{98.3} \\
EVL\includegraphics[width=0.25cm,height=0.25cm]{fig/frozen.pdf}~\cite{lin2022evl} & CLIP-400M  & ViT-L & $32\times336^2$ & $6.07\times1\times3$ & 363 & \cmark & 87.7 & - \\
Text4Vis\includegraphics[width=0.25cm,height=0.25cm]{fig/frozen.pdf}~\cite{wu2023revisiting} & CLIP-400M  & ViT-L & $32\times336^2$ & $3.83\times1\times3$ & 231 & \cmark & 87.8 & 97.6 \\
UnifromerV2\includegraphics[width=0.25cm,height=0.25cm]{fig/frozen.pdf}~\cite{li2022uniformerv2} & \scriptsize CLIP-400M+K710  & ViT-L & $32\times336^2$ & $6.27\times1\times3$& 354&\cmark  & 88.8 & 98.1 \\
\cellcolor{lightgrey}\textbf{DiST}$_{\gamma\text{=2}}$\includegraphics[width=0.25cm,height=0.25cm]{fig/frozen.pdf} &\cellcolor{lightgrey}CLIP-400M&\cellcolor{lightgrey}ViT-L &\cellcolor{lightgrey}$8\times336^2$ &\cellcolor{lightgrey}$1.66\times1\times3$ &\cellcolor{lightgrey}343&\cellcolor{lightgrey}\cmark&\cellcolor{lightgrey}87.2&\cellcolor{lightgrey}97.6 \\
\cellcolor{lightgrey}\textbf{DiST}$_{\gamma\text{=2}}$\includegraphics[width=0.25cm,height=0.25cm]{fig/frozen.pdf} &\cellcolor{lightgrey}CLIP-400M&\cellcolor{lightgrey}ViT-L &\cellcolor{lightgrey}$16\times336^2$ &\cellcolor{lightgrey}$3.32\times1\times3$ &\cellcolor{lightgrey}343&\cellcolor{lightgrey}\cmark&\cellcolor{lightgrey}87.9& \cellcolor{lightgrey}98.0 \\
\cellcolor{lightgrey}\textbf{DiST}$_{\gamma\text{=2}}$\includegraphics[width=0.25cm,height=0.25cm]{fig/frozen.pdf} &\cellcolor{lightgrey}CLIP-400M&\cellcolor{lightgrey}ViT-L &\cellcolor{lightgrey}$32\times336^2$ &\cellcolor{lightgrey}$6.64\times1\times3$ &\cellcolor{lightgrey}343&\cellcolor{lightgrey}\cmark&\cellcolor{lightgrey}88.5&\cellcolor{lightgrey}98.2 \\
\cellcolor{lightgrey}\textbf{DiST}$_{\gamma\text{=2}}$\includegraphics[width=0.25cm,height=0.25cm]{fig/frozen.pdf} &\cellcolor{lightgrey}\scriptsize CLIP-400M+K710&\cellcolor{lightgrey}ViT-L &\cellcolor{lightgrey}$8\times336^2$ &\cellcolor{lightgrey}$1.66\times1\times3$&\cellcolor{lightgrey}343&\cellcolor{lightgrey}\cmark&\cellcolor{lightgrey}88.1&\cellcolor{lightgrey}97.9 \\
\cellcolor{lightgrey}\textbf{DiST}$_{\gamma\text{=2}}$\includegraphics[width=0.25cm,height=0.25cm]{fig/frozen.pdf} &\cellcolor{lightgrey}\scriptsize CLIP-400M+K710&\cellcolor{lightgrey}ViT-L &\cellcolor{lightgrey}$16\times336^2$ &\cellcolor{lightgrey}$3.32\times1\times3$&\cellcolor{lightgrey}343&\cellcolor{lightgrey}\cmark&\cellcolor{lightgrey}88.9&\cellcolor{lightgrey}98.2 \\
\cellcolor{lightgrey}\textbf{DiST}$_{\gamma\text{=2}}$\includegraphics[width=0.25cm,height=0.25cm]{fig/frozen.pdf} &\cellcolor{lightgrey}\scriptsize CLIP-400M+K710&\cellcolor{lightgrey}ViT-L &\cellcolor{lightgrey}$32\times336^2$ &\cellcolor{lightgrey}$6.64\times1\times3$&\cellcolor{lightgrey}343&\cellcolor{lightgrey}\cmark&\cellcolor{lightgrey}\textbf{89.7}&\cellcolor{lightgrey}\textbf{98.5} \\

\end{tabular}
    \caption{Comparison with state-of-the-arts on Kinetics-400. }
    \label{tab:sota_k400_appendix}
\end{table*}

\noindent
\textbf{More results on Something-Something V2~\cite{goyal2017ssv2} and Kinetics-400~\cite{kay2017k400}.}
Here, we supplement more results with different frames and resolutions on two datasets in Tab.~\ref{tab:sota_ssv2_appendix} and Tab.~\ref{tab:sota_k400_appendix} for reference. 
From the two tables, we can draw the following conclusions:
(i) The more temporal details brought by more frames is highly effective for both SSV2 and K400, regardless of model size and pre-training datasets used.
For example, increasing the frames from 8 to 32 can result in consistent performance improvements of around 2.0\% on SSV2 and around 1.5\% on K400. 
This is because SSV2 relies more on temporal details, while K400 depends more on spatial information.
(ii) Increasing the resolution from 224 to 336 has a relatively small impact on performance, typically around 0.3\%, on both datasets, which is consistent with existing literature~\cite{lin2022evl}. 
Interestingly, using higher resolutions is apparently less effective than increasing the size of the pre-training datasets or models. 
Therefore, in future research, we will further explore the potential of video models from both the model and data scale perspectives.

\section{Implementation details}
\label{sec:impl_details}
Tab.~\ref{tab:impl_detail} summarizes the fine-tuning configurations across multiple datasets and models. In almost all the different datasets and models, the configurations are shared, which demonstrates the extensive adaptability of our proposed DiST.

\begin{table*}[ht]
    \centering
    \begin{tabular}{l|cccccc}
        dataset & \multicolumn{2}{c}{Kinetics-400}  & \multicolumn{2}{c}{Something-Something V2} & \multicolumn{2}{c}{Epic-Kitchens 100}\\
        backbone & ViT-B & ViT-L & ViT-B & ViT-L & ViT-B & ViT-L \\
        \shline
        \makecell{$\alpha$ (dim. of integration branch)} & $1/2$ (384) &$3/8$ (384)& $1/2$ (384) &$3/8$ (384) & $1/2$ (384) &$3/8$ (384) \\
        $\beta$ (dim. of temporal encoder) & $1/8$ (96) &$3/32$ (96)& $1/8$ (96) &$3/32$ (96) & $1/8$ (96) &$3/32$ (96) \\
        $\gamma$ (frames of temporal encoder) & \multicolumn{6}{c}{$2$}\\
        \hline
        optimizer & \multicolumn{6}{c}{AdamW, learning rate=3.2e-4, weight decay=1e-4, betas=[0.9, 0.999]} \\
        batch size & \multicolumn{6}{c}{256} \\
        training epochs & \multicolumn{6}{c}{36} \\
        warmup epochs & \multicolumn{6}{c}{6} \\
        training crop scale & \multicolumn{6}{c}{[0.4, 1.0]} \\
        training crop size & \multicolumn{6}{c}{224} \\
        frame sampling rate & \multicolumn{6}{c}{TSN~\cite{wang2016tsn} uniform sampling} \\
        mirror & \multicolumn{6}{c}{\cmark} \\
        RandAugment~\cite{cubuk2020randaugment} & \multicolumn{6}{c}{\xmark} \\
        MixUp~\cite{zhang2017mixup} & \multicolumn{6}{c}{0.8} \\
        CutMix~\cite{yun2019cutmix} & \multicolumn{6}{c}{1.0} \\
        testing views & \multicolumn{6}{c}{3 temporal $\times$ 1 spatial}\\
    \end{tabular}
    \caption{{\textbf{Configurations for Kinetics-400, Something-Something V2 and Epic-Kitchens 100.}}}
    \label{tab:impl_detail}
\end{table*}

%% file: main.bbl
\begin{thebibliography}{10}\itemsep=-1pt

\bibitem{arnab2021vivit}
Anurag Arnab, Mostafa Dehghani, Georg Heigold, Chen Sun, Mario Lu{\v{c}}i{\'c}, and Cordelia Schmid.
\newblock Vivit: A video vision transformer.
\newblock In {\em ICCV}, pages 6836--6846, 2021.

\bibitem{bahng2022visualprompt}
Hyojin Bahng, Ali Jahanian, Swami Sankaranarayanan, and Phillip Isola.
\newblock Visual prompting: Modifying pixel space to adapt pre-trained models.
\newblock {\em arXiv preprint arXiv:2203.17274}, 2022.

\bibitem{bertasius2021timesformer}
Gedas Bertasius, Heng Wang, and Lorenzo Torresani.
\newblock Is space-time attention all you need for video understanding?
\newblock In {\em ICML}, volume~2, page~4, 2021.

\bibitem{bulat2021xvit}
Adrian Bulat, Juan~Manuel Perez~Rua, Swathikiran Sudhakaran, Brais Martinez, and Georgios Tzimiropoulos.
\newblock Space-time mixing attention for video transformer.
\newblock {\em NeurIPS}, 34:19594--19607, 2021.

\bibitem{carreira2018k600}
Joao Carreira, Eric Noland, Andras Banki-Horvath, Chloe Hillier, and Andrew Zisserman.
\newblock A short note about kinetics-600.
\newblock {\em arXiv preprint arXiv:1808.01340}, 2018.

\bibitem{carreira2019k700}
Joao Carreira, Eric Noland, Chloe Hillier, and Andrew Zisserman.
\newblock A short note on the kinetics-700 human action dataset.
\newblock {\em arXiv preprint arXiv:1907.06987}, 2019.

\bibitem{carreira2017i3d}
Joao Carreira and Andrew Zisserman.
\newblock Quo vadis, action recognition? a new model and the kinetics dataset.
\newblock In {\em CVPR}, pages 6299--6308, 2017.

\bibitem{chen2022adaptformer}
Shoufa Chen, Chongjian Ge, Zhan Tong, Jiangliu Wang, Yibing Song, Jue Wang, and Ping Luo.
\newblock Adaptformer: Adapting vision transformers for scalable visual recognition.
\newblock {\em arXiv preprint arXiv:2205.13535}, 2022.

\bibitem{cheng2021improving}
Xing Cheng, Hezheng Lin, Xiangyu Wu, Fan Yang, and Dong Shen.
\newblock Improving video-text retrieval by multi-stream corpus alignment and dual softmax loss.
\newblock {\em arXiv preprint arXiv:2109.04290}, 2021.

\bibitem{cubuk2020randaugment}
Ekin~D Cubuk, Barret Zoph, Jonathon Shlens, and Quoc~V Le.
\newblock Randaugment: Practical automated data augmentation with a reduced search space.
\newblock In {\em CVPR Workshops}, pages 702--703, 2020.

\bibitem{damen2020ek100}
Dima Damen, Hazel Doughty, Giovanni~Maria Farinella, Antonino Furnari, Evangelos Kazakos, Jian Ma, Davide Moltisanti, Jonathan Munro, Toby Perrett, Will Price, et~al.
\newblock Rescaling egocentric vision.
\newblock {\em arXiv preprint arXiv:2006.13256}, 2020.

\bibitem{diba2021vi2clr}
Ali Diba, Vivek Sharma, Reza Safdari, Dariush Lotfi, Saquib Sarfraz, Rainer Stiefelhagen, and Luc Van~Gool.
\newblock Vi2clr: Video and image for visual contrastive learning of representation.
\newblock In {\em ICCV}, pages 1502--1512, 2021.

\bibitem{dosovitskiy2020-vit}
Alexey Dosovitskiy, Lucas Beyer, Alexander Kolesnikov, Dirk Weissenborn, Xiaohua Zhai, Thomas Unterthiner, Mostafa Dehghani, Matthias Minderer, Georg Heigold, Sylvain Gelly, et~al.
\newblock An image is worth 16x16 words: Transformers for image recognition at scale.
\newblock In {\em ICLR}, 2020.

\bibitem{fan2021mvit}
Haoqi Fan, Bo Xiong, Karttikeya Mangalam, Yanghao Li, Zhicheng Yan, Jitendra Malik, and Christoph Feichtenhofer.
\newblock Multiscale vision transformers.
\newblock In {\em ICCV}, pages 6824--6835, 2021.

\bibitem{feichtenhofer2020x3d}
Christoph Feichtenhofer.
\newblock X3d: Expanding architectures for efficient video recognition.
\newblock In {\em CVPR}, pages 203--213, 2020.

\bibitem{feichtenhofermasked}
Christoph Feichtenhofer, Haoqi Fan, Yanghao Li, and Kaiming He.
\newblock Masked autoencoders as spatiotemporal learners.
\newblock {\em NeurIPS}, 2022.

\bibitem{feichtenhofer2019slowfast}
Christoph Feichtenhofer, Haoqi Fan, Jitendra Malik, and Kaiming He.
\newblock Slowfast networks for video recognition.
\newblock In {\em ICCV}, pages 6202--6211, 2019.

\bibitem{feichtenhofer2021largescale}
Christoph Feichtenhofer, Haoqi Fan, Bo Xiong, Ross Girshick, and Kaiming He.
\newblock A large-scale study on unsupervised spatiotemporal representation learning.
\newblock In {\em CVPR}, pages 3299--3309, 2021.

\bibitem{gao2021clipadapter}
Peng Gao, Shijie Geng, Renrui Zhang, Teli Ma, Rongyao Fang, Yongfeng Zhang, Hongsheng Li, and Yu Qiao.
\newblock Clip-adapter: Better vision-language models with feature adapters.
\newblock {\em arXiv preprint arXiv:2110.04544}, 2021.

\bibitem{goyal2017ssv2}
Raghav Goyal, Samira Ebrahimi~Kahou, Vincent Michalski, Joanna Materzynska, Susanne Westphal, Heuna Kim, Valentin Haenel, Ingo Fruend, Peter Yianilos, Moritz Mueller-Freitag, et~al.
\newblock The" something something" video database for learning and evaluating visual common sense.
\newblock In {\em ICCV}, pages 5842--5850, 2017.

\bibitem{huang2021tada}
Ziyuan Huang, Shiwei Zhang, Liang Pan, Zhiwu Qing, Mingqian Tang, Ziwei Liu, and Marcelo~H Ang~Jr.
\newblock Tada! temporally-adaptive convolutions for video understanding.
\newblock In {\em ICLR}, 2022.

\bibitem{huang2023tadav2}
Ziyuan Huang, Shiwei Zhang, Liang Pan, Zhiwu Qing, Yingya Zhang, Ziwei Liu, and Marcelo~H Ang~Jr.
\newblock Temporally-adaptive models for efficient video understanding.
\newblock {\em arXiv preprint arXiv:2308.05787}, 2023.

\bibitem{jenni2021time}
Simon Jenni and Hailin Jin.
\newblock Time-equivariant contrastive video representation learning.
\newblock In {\em ICCV}, pages 9970--9980, 2021.

\bibitem{jhuang2011hmdb51}
H Jhuang, H Garrote, E Poggio, T Serre, and T Hmdb.
\newblock A large video database for human motion recognition.
\newblock In {\em ICCV}, volume~4, page~6, 2011.

\bibitem{jia2021scaling}
Chao Jia, Yinfei Yang, Ye Xia, Yi-Ting Chen, Zarana Parekh, Hieu Pham, Quoc Le, Yun-Hsuan Sung, Zhen Li, and Tom Duerig.
\newblock Scaling up visual and vision-language representation learning with noisy text supervision.
\newblock In {\em ICML}, pages 4904--4916. PMLR, 2021.

\bibitem{jiang2019stm}
Boyuan Jiang, MengMeng Wang, Weihao Gan, Wei Wu, and Junjie Yan.
\newblock Stm: Spatiotemporal and motion encoding for action recognition.
\newblock In {\em ICCV}, pages 2000--2009, 2019.

\bibitem{ju2022promptingclipvid}
Chen Ju, Tengda Han, Kunhao Zheng, Ya Zhang, and Weidi Xie.
\newblock Prompting visual-language models for efficient video understanding.
\newblock In {\em ECCV}, pages 105--124. Springer, 2022.

\bibitem{ju2022prompting}
Chen Ju, Tengda Han, Kunhao Zheng, Ya Zhang, and Weidi Xie.
\newblock Prompting visual-language models for efficient video understanding.
\newblock In {\em ECCV}, pages 105--124. Springer, 2022.

\bibitem{karpathy2014twostream}
Andrej Karpathy, George Toderici, Sanketh Shetty, Thomas Leung, Rahul Sukthankar, and Li Fei-Fei.
\newblock Large-scale video classification with convolutional neural networks.
\newblock In {\em CVPR}, pages 1725--1732, 2014.

\bibitem{kay2017k400}
Will Kay, Joao Carreira, Karen Simonyan, Brian Zhang, Chloe Hillier, Sudheendra Vijayanarasimhan, Fabio Viola, Tim Green, Trevor Back, Paul Natsev, et~al.
\newblock The kinetics human action video dataset.
\newblock {\em arXiv preprint arXiv:1705.06950}, 2017.

\bibitem{cka}
Simon Kornblith, Mohammad Norouzi, Honglak Lee, and Geoffrey Hinton.
\newblock Similarity of neural network representations revisited.
\newblock In {\em ICML}, volume~97, pages 3519--3529. PMLR, 09--15 Jun 2019.

\bibitem{li2022uniformerv2}
Kunchang Li, Yali Wang, Yinan He, Yizhuo Li, Yi Wang, Limin Wang, and Yu Qiao.
\newblock Uniformerv2: Spatiotemporal learning by arming image vits with video uniformer.
\newblock {\em arXiv preprint arXiv:2211.09552}, 2022.

\bibitem{li2022uniformer}
Kunchang Li, Yali Wang, Junhao Zhang, Peng Gao, Guanglu Song, Yu Liu, Hongsheng Li, and Yu Qiao.
\newblock Uniformer: Unifying convolution and self-attention for visual recognition.
\newblock {\em TPAMI}, 2023.

\bibitem{li2021prefix}
Xiang~Lisa Li and Percy Liang.
\newblock Prefix-tuning: Optimizing continuous prompts for generation.
\newblock {\em arXiv preprint arXiv:2101.00190}, 2021.

\bibitem{li2022flip}
Yanghao Li, Haoqi Fan, Ronghang Hu, Christoph Feichtenhofer, and Kaiming He.
\newblock Scaling language-image pre-training via masking.
\newblock {\em arXiv preprint arXiv:2212.00794}, 2022.

\bibitem{li2020tea}
Yan Li, Bin Ji, Xintian Shi, Jianguo Zhang, Bin Kang, and Limin Wang.
\newblock Tea: Temporal excitation and aggregation for action recognition.
\newblock In {\em CVPR}, pages 909--918, 2020.

\bibitem{li2022mvitv2}
Yanghao Li, Chao-Yuan Wu, Haoqi Fan, Karttikeya Mangalam, Bo Xiong, Jitendra Malik, and Christoph Feichtenhofer.
\newblock Mvitv2: Improved multiscale vision transformers for classification and detection.
\newblock In {\em CVPR}, pages 4804--4814, 2022.

\bibitem{lian2022ssfada}
Dongze Lian, Daquan Zhou, Jiashi Feng, and Xinchao Wang.
\newblock Scaling \& shifting your features: A new baseline for efficient model tuning.
\newblock {\em arXiv preprint arXiv:2210.08823}, 2022.

\bibitem{lin2019tsm}
Ji Lin, Chuang Gan, and Song Han.
\newblock Tsm: Temporal shift module for efficient video understanding.
\newblock In {\em ICCV}, pages 7083--7093, 2019.

\bibitem{lin2022evl}
Ziyi Lin, Shijie Geng, Renrui Zhang, Peng Gao, Gerard de Melo, Xiaogang Wang, Jifeng Dai, Yu Qiao, and Hongsheng Li.
\newblock Frozen clip models are efficient video learners.
\newblock In {\em ECCV}, pages 388--404. Springer, 2022.

\bibitem{liu2022videoswin}
Ze Liu, Jia Ning, Yue Cao, Yixuan Wei, Zheng Zhang, Stephen Lin, and Han Hu.
\newblock Video swin transformer.
\newblock In {\em CVPR}, pages 3202--3211, 2022.

\bibitem{miech2020milnce}
Antoine Miech, Jean-Baptiste Alayrac, Lucas Smaira, Ivan Laptev, Josef Sivic, and Andrew Zisserman.
\newblock End-to-end learning of visual representations from uncurated instructional videos.
\newblock In {\em CVPR}, pages 9879--9889, 2020.

\bibitem{miech2019howto100m}
Antoine Miech, Dimitri Zhukov, Jean-Baptiste Alayrac, Makarand Tapaswi, Ivan Laptev, and Josef Sivic.
\newblock Howto100m: Learning a text-video embedding by watching hundred million narrated video clips.
\newblock In {\em ICCV}, pages 2630--2640, 2019.

\bibitem{neimark2021vtn}
Daniel Neimark, Omri Bar, Maya Zohar, and Dotan Asselmann.
\newblock Video transformer network.
\newblock In {\em ICCV}, pages 3163--3172, 2021.

\bibitem{ni2022expanding-xclip}
Bolin Ni, Houwen Peng, Minghao Chen, Songyang Zhang, Gaofeng Meng, Jianlong Fu, Shiming Xiang, and Haibin Ling.
\newblock Expanding language-image pretrained models for general video recognition.
\newblock In {\em ECCV}, pages 1--18. Springer, 2022.

\bibitem{pan2022st-adapter}
Junting Pan, Ziyi Lin, Xiatian Zhu, Jing Shao, and Hongsheng Li.
\newblock Parameter-efficient image-to-video transfer learning.
\newblock {\em arXiv e-prints}, pages arXiv--2206, 2022.

\bibitem{pan2021videomoco}
Tian Pan, Yibing Song, Tianyu Yang, Wenhao Jiang, and Wei Liu.
\newblock Videomoco: Contrastive video representation learning with temporally adversarial examples.
\newblock In {\em CVPR}, pages 11205--11214, 2021.

\bibitem{patrick2021motionformer}
Mandela Patrick, Dylan Campbell, Yuki Asano, Ishan Misra, Florian Metze, Christoph Feichtenhofer, Andrea Vedaldi, and Jo{\~a}o~F Henriques.
\newblock Keeping your eye on the ball: Trajectory attention in video transformers.
\newblock {\em NeurIPS}, 34:12493--12506, 2021.

\bibitem{pfeiffer2020adapterfusion}
Jonas Pfeiffer, Aishwarya Kamath, Andreas R{\"u}ckl{\'e}, Kyunghyun Cho, and Iryna Gurevych.
\newblock Adapterfusion: Non-destructive task composition for transfer learning.
\newblock {\em arXiv preprint arXiv:2005.00247}, 2020.

\bibitem{qian2021cvrl}
Rui Qian, Tianjian Meng, Boqing Gong, Ming-Hsuan Yang, Huisheng Wang, Serge Belongie, and Yin Cui.
\newblock Spatiotemporal contrastive video representation learning.
\newblock In {\em CVPR}, pages 6964--6974, 2021.

\bibitem{qiu2017p3d}
Zhaofan Qiu, Ting Yao, and Tao Mei.
\newblock Learning spatio-temporal representation with pseudo-3d residual networks.
\newblock In {\em ICCV}, pages 5533--5541, 2017.

\bibitem{qiu2019learning}
Zhaofan Qiu, Ting Yao, Chong-Wah Ngo, Xinmei Tian, and Tao Mei.
\newblock Learning spatio-temporal representation with local and global diffusion.
\newblock In {\em CVPR}, pages 12056--12065, 2019.

\bibitem{radford2021clip}
Alec Radford, Jong~Wook Kim, Chris Hallacy, Aditya Ramesh, Gabriel Goh, Sandhini Agarwal, Girish Sastry, Amanda Askell, Pamela Mishkin, Jack Clark, et~al.
\newblock Learning transferable visual models from natural language supervision.
\newblock In {\em ICML}, pages 8748--8763. PMLR, 2021.

\bibitem{ryoo2021tokenlearner}
Michael Ryoo, AJ Piergiovanni, Anurag Arnab, Mostafa Dehghani, and Anelia Angelova.
\newblock Tokenlearner: Adaptive space-time tokenization for videos.
\newblock {\em NeurIPS}, 34:12786--12797, 2021.

\bibitem{simonyan2014twostream}
Karen Simonyan and Andrew Zisserman.
\newblock Two-stream convolutional networks for action recognition in videos.
\newblock {\em NeurIPS}, 27, 2014.

\bibitem{soomro2012ucf101}
Khurram Soomro, Amir~Roshan Zamir, and Mubarak Shah.
\newblock Ucf101: A dataset of 101 human actions classes from videos in the wild.
\newblock {\em arXiv preprint arXiv:1212.0402}, 2012.

\bibitem{sun2019videobert}
Chen Sun, Austin Myers, Carl Vondrick, Kevin Murphy, and Cordelia Schmid.
\newblock Videobert: A joint model for video and language representation learning.
\newblock In {\em ICCV}, pages 7464--7473, 2019.

\bibitem{sung2022vl-adapter}
Yi-Lin Sung, Jaemin Cho, and Mohit Bansal.
\newblock Vl-adapter: Parameter-efficient transfer learning for vision-and-language tasks.
\newblock In {\em CVPR}, pages 5227--5237, 2022.

\bibitem{tong2022videomaenju}
Zhan Tong, Yibing Song, Jue Wang, and Limin Wang.
\newblock Videomae: Masked autoencoders are data-efficient learners for self-supervised video pre-training.
\newblock {\em NeurIPS}, 2022.

\bibitem{tran2015c3d}
Du Tran, Lubomir Bourdev, Rob Fergus, Lorenzo Torresani, and Manohar Paluri.
\newblock Learning spatiotemporal features with 3d convolutional networks.
\newblock In {\em ICCV}, pages 4489--4497, 2015.

\bibitem{tran2019csn}
Du Tran, Heng Wang, Lorenzo Torresani, and Matt Feiszli.
\newblock Video classification with channel-separated convolutional networks.
\newblock In {\em ICCV}, pages 5552--5561, 2019.

\bibitem{tran2018r21d}
Du Tran, Heng Wang, Lorenzo Torresani, Jamie Ray, Yann LeCun, and Manohar Paluri.
\newblock A closer look at spatiotemporal convolutions for action recognition.
\newblock In {\em CVPR}, pages 6450--6459, 2018.

\bibitem{wang2020video}
Heng Wang, Du Tran, Lorenzo Torresani, and Matt Feiszli.
\newblock Video modeling with correlation networks.
\newblock In {\em CVPR}, pages 352--361, 2020.

\bibitem{wang2018artnet}
Limin Wang, Wei Li, Wen Li, and Luc Van~Gool.
\newblock Appearance-and-relation networks for video classification.
\newblock In {\em CVPR}, pages 1430--1439, 2018.

\bibitem{wang2021tdn}
Limin Wang, Zhan Tong, Bin Ji, and Gangshan Wu.
\newblock Tdn: Temporal difference networks for efficient action recognition.
\newblock In {\em CVPR}, pages 1895--1904, 2021.

\bibitem{wang2016tsn}
Limin Wang, Yuanjun Xiong, Zhe Wang, Yu Qiao, Dahua Lin, Xiaoou Tang, and Luc~Van Gool.
\newblock Temporal segment networks: Towards good practices for deep action recognition.
\newblock In {\em ECCV}, pages 20--36. Springer, 2016.

\bibitem{wang2018tsn}
Limin Wang, Yuanjun Xiong, Zhe Wang, Yu Qiao, Dahua Lin, Xiaoou Tang, and Luc Van~Gool.
\newblock Temporal segment networks for action recognition in videos.
\newblock {\em TPAMI}, 41(11):2740--2755, 2018.

\bibitem{wang2021actionclip}
Mengmeng Wang, Jiazheng Xing, and Yong Liu.
\newblock Actionclip: A new paradigm for video action recognition.
\newblock {\em arXiv preprint arXiv:2109.08472}, 2021.

\bibitem{wang2022bevt}
Rui Wang, Dongdong Chen, Zuxuan Wu, Yinpeng Chen, Xiyang Dai, Mengchen Liu, Yu-Gang Jiang, Luowei Zhou, and Lu Yuan.
\newblock Bevt: Bert pretraining of video transformers.
\newblock In {\em CVPR}, pages 14733--14743, 2022.

\bibitem{wei2022maskedfeat}
Chen Wei, Haoqi Fan, Saining Xie, Chao-Yuan Wu, Alan Yuille, and Christoph Feichtenhofer.
\newblock Masked feature prediction for self-supervised visual pre-training.
\newblock In {\em CVPR}, pages 14668--14678, 2022.

\bibitem{wu2023revisiting}
Wenhao Wu, Zhun Sun, and Wanli Ouyang.
\newblock Revisiting classifier: Transferring vision-language models for video recognition.
\newblock In {\em AAAI}, volume~37, pages 2847--2855, 2023.

\bibitem{wu2023bidirectional}
Wenhao Wu, Xiaohan Wang, Haipeng Luo, Jingdong Wang, Yi Yang, and Wanli Ouyang.
\newblock Bidirectional cross-modal knowledge exploration for video recognition with pre-trained vision-language models.
\newblock In {\em CVPR}, pages 6620--6630, 2023.

\bibitem{yan2022multiview}
Shen Yan, Xuehan Xiong, Anurag Arnab, Zhichao Lu, Mi Zhang, Chen Sun, and Cordelia Schmid.
\newblock Multiview transformers for video recognition.
\newblock In {\em CVPR}, pages 3333--3343, 2022.

\bibitem{yang2020tpn}
Ceyuan Yang, Yinghao Xu, Jianping Shi, Bo Dai, and Bolei Zhou.
\newblock Temporal pyramid network for action recognition.
\newblock In {\em CVPR}, pages 591--600, 2020.

\bibitem{yu2022coca}
Jiahui Yu, Zirui Wang, Vijay Vasudevan, Legg Yeung, Mojtaba Seyedhosseini, and Yonghui Wu.
\newblock Coca: Contrastive captioners are image-text foundation models.
\newblock {\em arXiv preprint arXiv:2205.01917}, 2022.

\bibitem{yuan2021florence}
Lu Yuan, Dongdong Chen, Yi-Ling Chen, Noel Codella, Xiyang Dai, Jianfeng Gao, Houdong Hu, Xuedong Huang, Boxin Li, Chunyuan Li, et~al.
\newblock Florence: A new foundation model for computer vision.
\newblock {\em arXiv preprint arXiv:2111.11432}, 2021.

\bibitem{yun2019cutmix}
Sangdoo Yun, Dongyoon Han, Seong~Joon Oh, Sanghyuk Chun, Junsuk Choe, and Youngjoon Yoo.
\newblock Cutmix: Regularization strategy to train strong classifiers with localizable features.
\newblock In {\em ICCV}, pages 6023--6032, 2019.

\bibitem{zhang2017mixup}
Hongyi Zhang, Moustapha Cisse, Yann~N Dauphin, and David Lopez-Paz.
\newblock mixup: Beyond empirical risk minimization.
\newblock {\em arXiv preprint arXiv:1710.09412}, 2017.

\bibitem{zhang2021vt-clip}
Renrui Zhang, Longtian Qiu, Wei Zhang, and Ziyao Zeng.
\newblock Vt-clip: Enhancing vision-language models with visual-guided texts.
\newblock {\em arXiv preprint arXiv:2112.02399}, 2021.

\bibitem{zhou2022cocop}
Kaiyang Zhou, Jingkang Yang, Chen~Change Loy, and Ziwei Liu.
\newblock Conditional prompt learning for vision-language models.
\newblock In {\em CVPR}, pages 16816--16825, 2022.

\bibitem{zhou2022coop}
Kaiyang Zhou, Jingkang Yang, Chen~Change Loy, and Ziwei Liu.
\newblock Learning to prompt for vision-language models.
\newblock {\em IJCV}, 130(9):2337--2348, 2022.

\bibitem{zhu2020actbert}
Linchao Zhu and Yi Yang.
\newblock Actbert: Learning global-local video-text representations.
\newblock In {\em CVPR}, pages 8746--8755, 2020.

\end{thebibliography}
